\documentclass{isprs} 
\usepackage{subfigure}
\usepackage{setspace}
\usepackage{geometry} 
\usepackage{epstopdf}
\usepackage[labelsep=period]{caption}  
\usepackage[british]{babel} 
\usepackage[hang]{footmisc}
\usepackage{natbib}

\usepackage[T1]{fontenc}
\usepackage{siunitx}
\usepackage{array,multirow}
\usepackage{colortbl}
\usepackage{xcolor}
\usepackage{arydshln}
\usepackage{caption}
\usepackage[export]{adjustbox}
\usepackage{subcaption}

\usepackage{todonotes}

\begin{document}

\title{A Critical Synthesis of Uncertainty Quantification and Foundation Models\\ in Monocular Depth Estimation}
\date{}
\author{S. Landgraf\textsuperscript{1,}\thanks{Corresponding Author.} , R. Qin\textsuperscript{2 }, M. Ulrich\textsuperscript{1 }}

\address{\textsuperscript{1 }Institute of Photogrammetry and Remote Sensing (IPF), Karlsruhe Institute of Technology (KIT), Germany -\\ (steven.landgraf, markus.ulrich)@kit.edu\\ \textsuperscript{2 }The Ohio State University, Columbus, Ohio, United States -\\
{qin.324}@osu.edu}

\abstract{
While recent foundation models have enabled significant breakthroughs in monocular depth estimation, a clear path towards safe and reliable deployment in the real-world remains elusive. Metric depth estimation, which involves predicting absolute distances, poses particular challenges, as even the most advanced foundation models remain prone to critical errors. Since quantifying the uncertainty has emerged as a promising endeavor to address these limitations and enable trustworthy deployment, we fuse five different uncertainty quantification methods with the current state-of-the-art DepthAnythingV2 foundation model. To cover a wide range of metric depth domains, we evaluate their performance on four diverse datasets. Our findings identify fine-tuning with the Gaussian Negative Log-Likelihood Loss (GNLL) as a particularly promising approach, offering reliable uncertainty estimates while maintaining predictive performance and computational efficiency on par with the baseline, encompassing both training and inference time. By fusing uncertainty quantification and foundation models within the context of monocular depth estimation, this paper lays a critical foundation for future research aimed at improving not only model performance but also its explainability. Extending this critical synthesis of uncertainty quantification and foundation models into other crucial tasks, such as semantic segmentation and pose estimation, presents exciting opportunities for safer and more reliable machine vision systems.
}


\maketitle

\section{Introduction}\label{INTRODUCTION}
\sloppy
Monocular depth estimation (MDE) received significant attention in recent years due to its crucial role in various downstream tasks ranging from autonomous driving \citep{xue2020toward,xiang2022visual} and robotics \citep{dong2022towards,roussel2019monocular} to AI-generated content such as images \citep{zhang2023adding}, videos \citep{liew2023magicedit}, and 3D scenes \citep{xu2023neurallift,shahbazi2024inserf,shriram2024realmdreamer}. At its core, MDE aims to transform a single image into a depth map by regressing range values for each pixel, all without exploiting direct range or stereo measurements. Theoretically, MDE is a geometrically ill-posed problem that is fundamentally ambiguous and can only be solved with the help of prior knowledge about object shapes, sizes, scene layouts, and occlusion patterns. This inherent requirement for scene understanding perfectly aligns MDE with deep learning approaches, which have proven proficient in encoding potent priors \citep{bhat2023zoedepth,Piccinelli_2024_CVPR,yang2024depth_1,yang2024depth_2}. These models benefit from extreme scaling, i.e., training on massive datasets and increasing model size, which facilitates the emergence of high-level visual scene understanding.

Based on these findings, a plethora of models have been proposed to address the challenges of MDE, with recent state-of-the-art solutions often leveraging large vision transformers trained on internet-scale data \citep{chen2016single,chen2020oasis,li2018megadepth,ranftl2020towards,bhat2023zoedepth,Piccinelli_2024_CVPR,yang2024depth_1,yang2024depth_2}, yielding foundation models capable of generalizing to a wide range of applications and scenes. A particularly challenging yet crucial application in fields such as robotics \citep{dong2022towards,roussel2019monocular}, augmented reality \citep{kalia2019real}, and autonomous driving \citep{xue2020toward,xiang2022visual} is the estimation of absolute distances in real-world units (e.g., meters), commonly referred to as metric depth estimation. This task is especially difficult due to inherent metric ambiguities caused by different camera models and scene variations. Fortunately, these foundation models can successfully be fine-tuned in the respective domain \citep{bhat2023zoedepth,Piccinelli_2024_CVPR,yang2024depth_1,yang2024depth_2} to determine exceptionally accurate metric depths.  

However, the strong performance of foundation models on common benchmarks \citep{silberman2012indoor,geiger2012we,cordts2016cityscapes,song2015sun} can lead to naive deployment, potentially overlooking their limitations. This can be particularly detrimental in safety-critical applications where errors can have serious consequences. There are multiple challenges associated with real-world deployment of deep learning models, including the lack of transparency due to the "black box" character of end-to-end systems \citep{roy2019bayesian,gawlikowski2022SurveyUncertainty}, the inability to distinguish between in-domain and out-of-domain samples \citep{lee2022trust,lee2017training}, the tendency to be overconfident \citep{guo2017calibration}, and the sensitivity to adversarial attacks \citep{rawat2017harnessing,serban2018adversarial,smith2018understanding}. 

\begin{figure}[t!]
\centering
\subfigure[Input Image]{\includegraphics[width=0.23\textwidth]{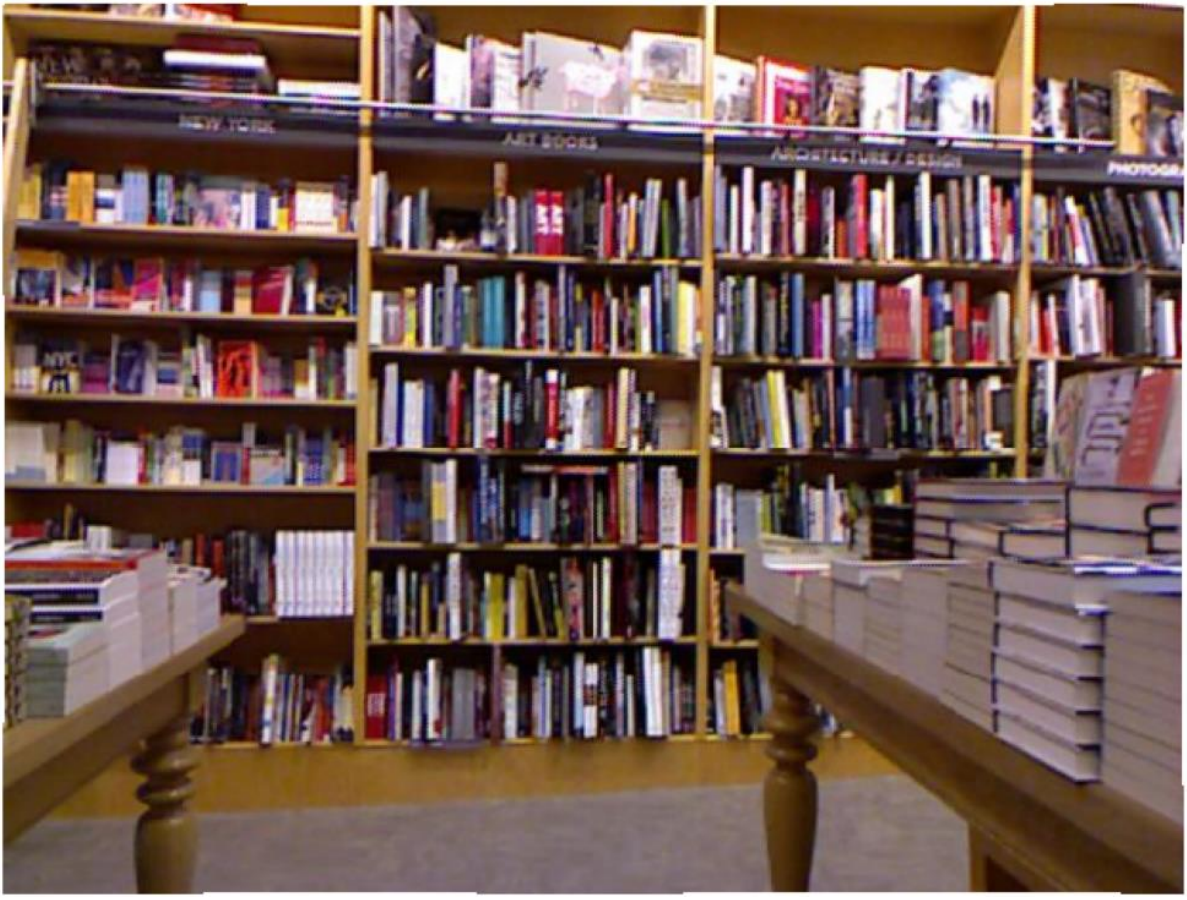}}
\subfigure[Prediction]{\includegraphics[width=0.23\textwidth]{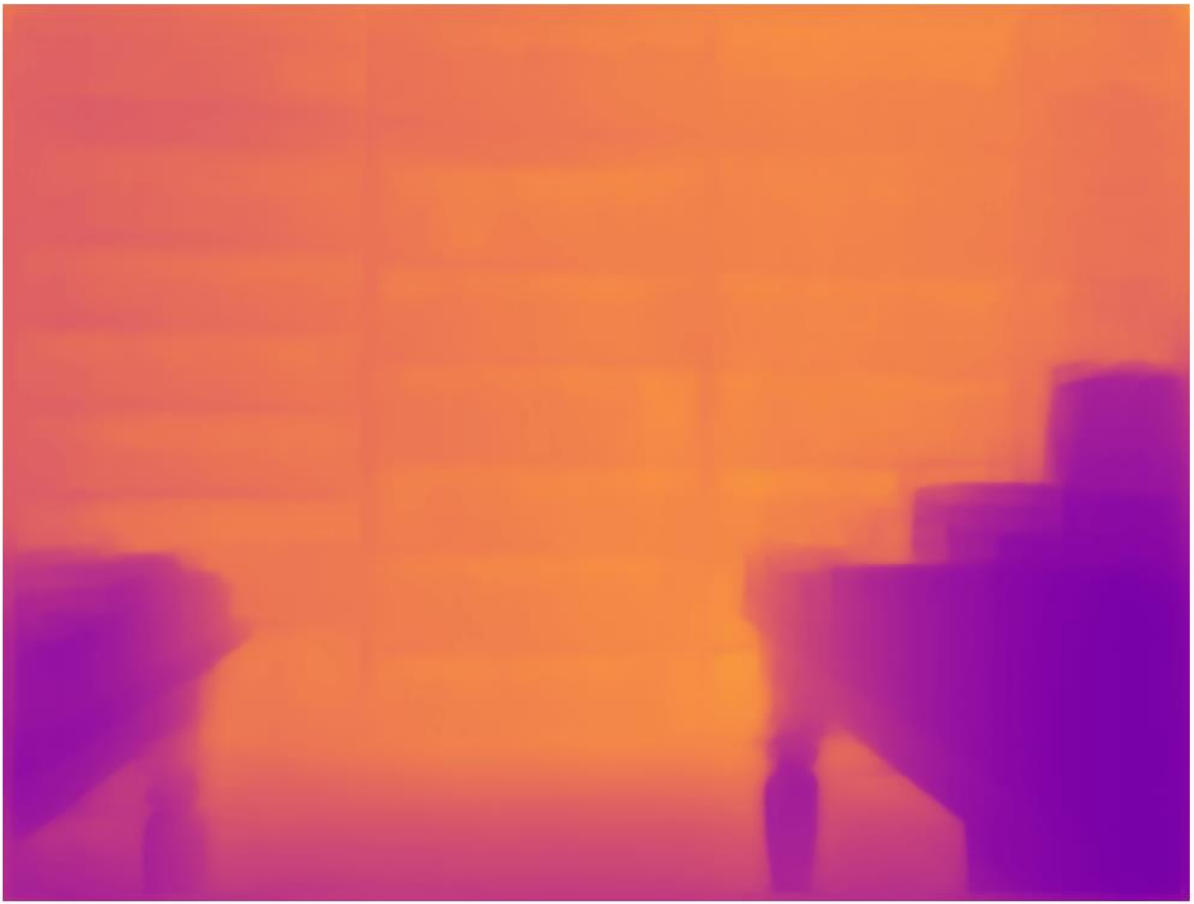}}
\subfigure[Binary Accuracy Map]{\includegraphics[width=0.23\textwidth]{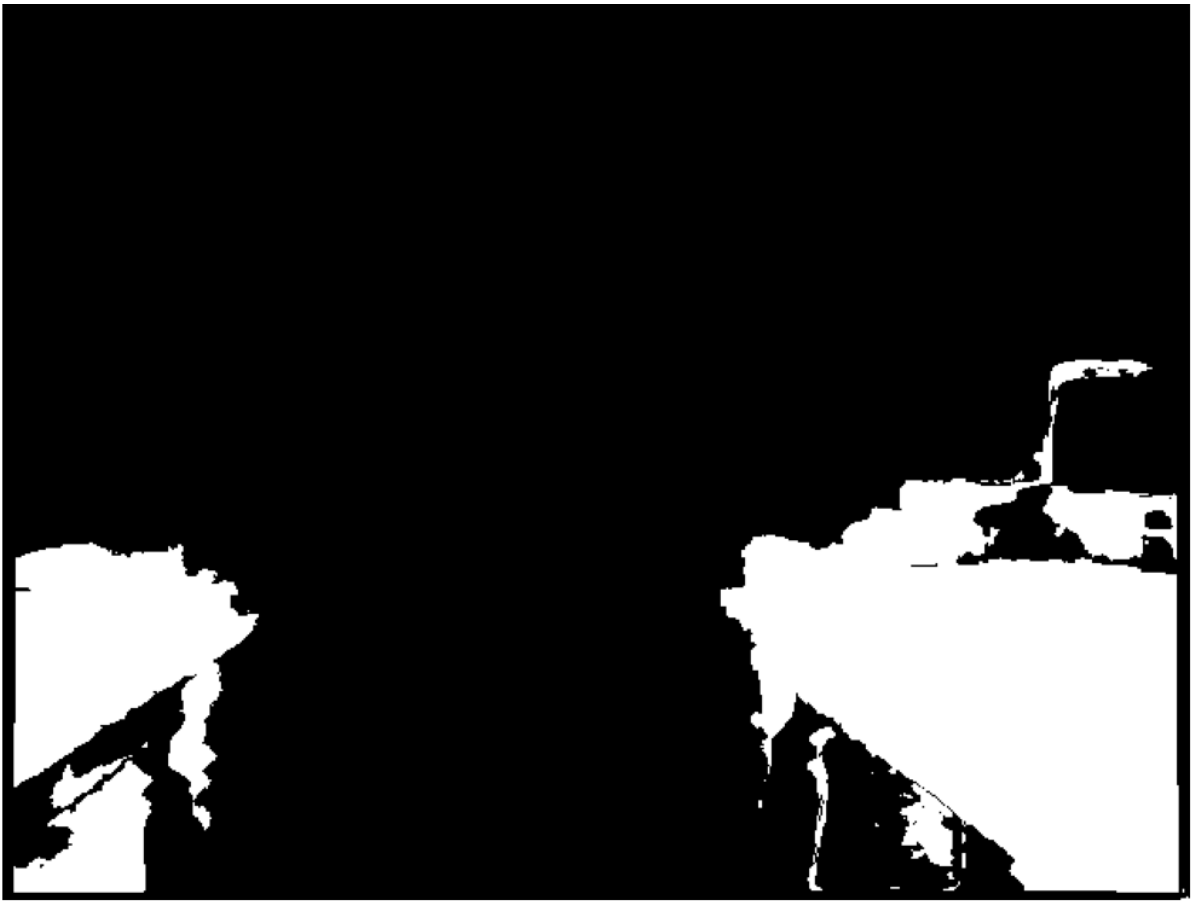}}
\subfigure[Uncertainty]{\includegraphics[width=0.23\textwidth]{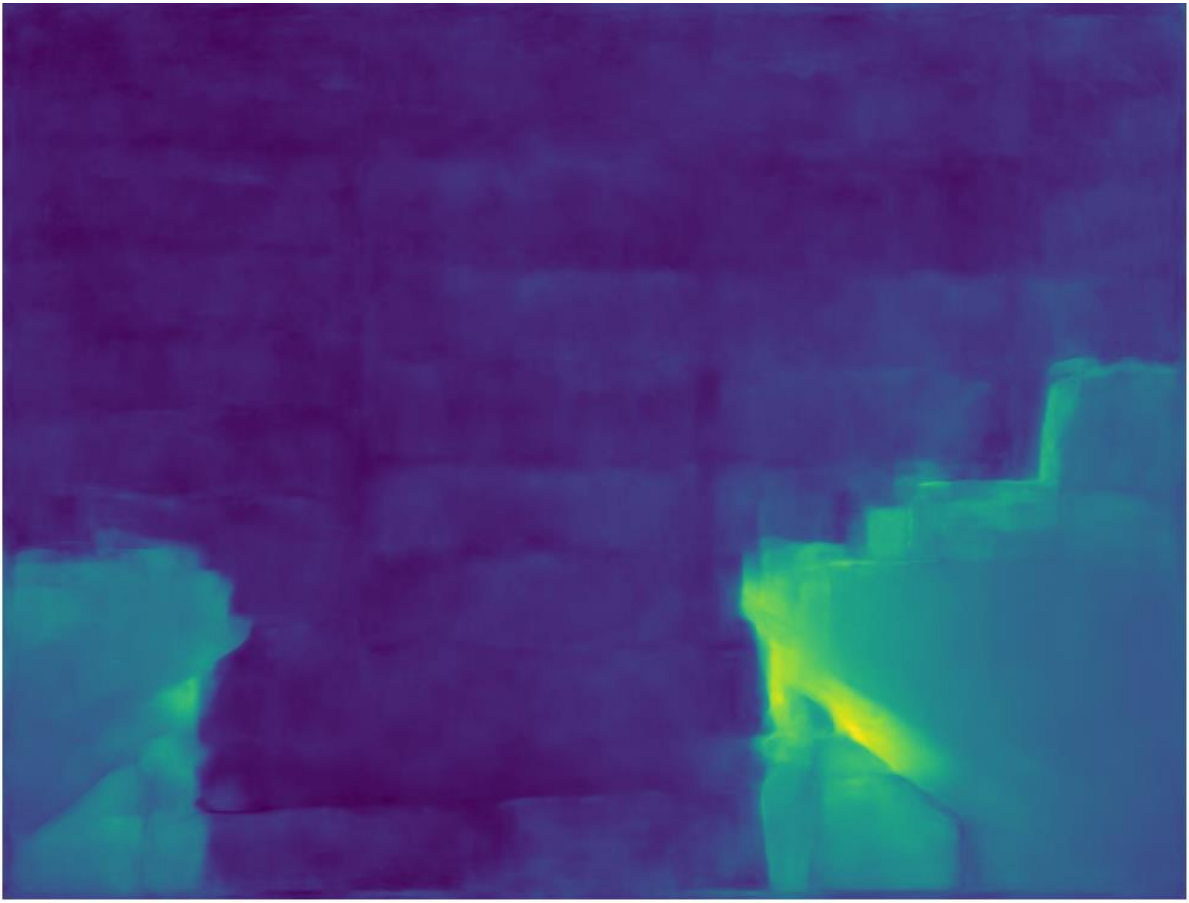}}
\caption{Qualitative example of a fine-tuned DepthAnythingV2 for metric monocular depth estimation on the NYUv2 dataset \protect\citep{silberman2012indoor}, using a ViT-S encoder and Monte Carlo Dropout for an additional uncertainty estimate. The binary accuracy map is based on the $\delta_1$ error. The strong correlation between erroneous predictions and high uncertainties highlights the potential of integrating uncertainty quantification (UQ) methods with foundation models for MDE.}
\label{fig: intro}
\end{figure}

To mitigate these risks, recent research has advocated for the quantification of uncertainty in deep learning models \citep{landgraf2024dudes,leibig2017LeveragingUncertainty,lee2018TrainingConfidencecalibrated,mukhoti2018evaluating,mukhoti2023deep,landgraf2024uncertainty,loquercio2020general}, particularly in scenarios where their deployment could have real-world implications. A number of promising methods for uncertainty quantification (UQ) have been developed, yet, surprisingly, there has been little attention on the integration of UQ with powerful MDE foundation models. As illustrated in Figure \ref{fig: intro}, even state-of-the-art foundation models are not immune to inaccuracies. However, if we can leverage UQ to correlate high uncertainties with erroneous predictions, it opens up the possibility of safer deployment of these models in real-world applications. 


To bridge the gap between ground-breaking results in research and safe, reliable deployment in real-world applications, we investigate multiple UQ methods in combination with MDE foundation models. We specifically focus on combining the state-of-the-art DepthAnythingV2 foundation model \citep{yang2024depth_2} with five different UQ methods to enable pixel-wise uncertainty measures for metric depth estimation:
\begin{enumerate}
    \item \textbf{Learned Confidence (LC)} \citep{wan2018confnet}: Confidences, interpreted as uncertainties, are learned by extending the primary objective function with an additional loss term.
    \item \textbf{Gaussian Negative Log-Likelihood (GNLL)} \citep{nix1994estimating}: Predictions are treated as samples from a Gaussian distribution, with the network outputting both a predictive mean and its corresponding variance, which is learned implicitly through minimizing the Gaussian Negative Log-Likelihood. 
    \item \textbf{Monte Carlo Dropout (MCD)} \citep{gal2016DropoutBayesian}: Dropout layers remain active during inference, sampling from the posterior distribution to estimate a predictive mean and variance.
    \item \textbf{Sub-Ensembles (SE)} \citep{valdenegro2023sub}: A Deep Ensemble \citep{lakshminarayanan2017SimpleScalable} is approximated by multiplying a subset of the model's layers instead of using the full model.
    \item \textbf{Test-Time Augmentation (TTA)} \citep{ayhan2018test}: Perturbations applied to inputs during inference produce unique samples, enabling computation of a predictive mean and corresponding variance.
\end{enumerate}

We conduct extensive evaluations across four diverse datasets:
\begin{enumerate} 
    \item \textbf{NYUv2} \citep{silberman2012indoor}: Low-resolution indoor scenes.
    \item \textbf{Cityscapes} \citep{cordts2016cityscapes}: High-resolution outdoor urban scenes. 
    \item \textbf{HOPE} \citep{tyree2022hope}: Synthetic training images and real-world testing images of household objects.
    \item \textbf{UseGeo} \citep{nex2024usegeo}: High-resolution aerial images.
\end{enumerate}

This wide selection ensures coverage of a broad spectrum of real-world applications. Additionally, we examine all publicly available pre-trained encoders of the DepthAnythingV2 model, comprehensively revealing insights into the performance of models across various sizes and configurations. 

We believe that our analysis will contribute to a deeper understanding of the limitations of these powerful foundation models, enhancing transparency and reliability in MDE. By highlighting often overlooked challenges, such as the lack of explainability, our work paves the way for future research that not only seeks to improve performance but also aims to teach these models to know what they don't know.

\section{Related Work}\label{RELATED_WORK}
\subsection{Monocular Depth Estimation}
\textbf{Foundations.} Monocular Depth Estimation (MDE) is a dense regression task that aims to predict a depth value for each pixel in a given input image. The pioneering work of \citet{eigen2014depth} laid the foundation for MDE by directly predicting depth using a multi-scale neural network. This seminal approach demonstrated that convolutional neural networks could effectively learn spatial hierarchies and capture depth cues from monocular images, thus inspiring a plethora of subsequent methods \citep{ming2021deep,masoumian2022monocular,khan2020deep,arampatzakis2023monocular}. While most introduce novel architectures or loss functions, \citet{fu2018deep} reformulate depth estimation as an ordinal regression problem through discretization of the depth ranges. Another innovative approach by \citet{yuan2022neural} incorporates neural conditional random fields to model contextual dependencies, further refining depth predictions. Besides, \citet{patil2022p3depth} impose geometric constraints based on piecewise planarity priors. 

\textbf{Vision Transformers.} Naturally, the rise of vision transformers \citep{dosovitskiy2020image} has also significantly impacted the field of MDE. These models employ the self-attention mechanism of the transformer to aggregate depth information across a more extensive field of view to capture long-range dependencies and global context, leading to more accurate and consistent depth maps. Inherently, multiple approaches have successfully adapted vision transformers to MDE \citep{agarwal2022depthformer,li2023depthformer,yang2021transformer,aich2021bidirectional,zhao2022monovit,ning2023all,ranftl2021vision,eftekhar2021omnidata,yin2021learning,piccinelli2023idisc,ke2024repurposing}. 

\textbf{Hybrid Approaches.} Beyond more traditional regression-based methods, some works creatively treat MDE as a combined regression-classification task. By discretizing the depth range into bins, these methods simplify the learning task improve performance in some cases. Notable examples include AdaBins \citep{bhat2021adabins}, BinsFormer \citep{li2024binsformer}, and LocalBins \citep{bhat2023zoedepth}. 

\textbf{Generative Models.} A more recent trend in MDE includes repurposing generative models such as diffusion models \citep{duan2023diffusiondepth,ji2023ddp,saxena2024surprising,saxena2023monocular,ke2024repurposing,patni2024ecodepth}, effectively building on its predecessor, generative adversarial networks \citep{cs2018monocular,aleotti2018generative}.

\textbf{Depth in the Wild.} Estimating depth "in the wild" has become an increasingly important area of research in MDE. It refers to the challenge of predicting accurate depth estimates in unconstrained environments, where lighting, scene structure, and camera parameters vary significantly. With the increasing availability of compute resources, researchers have found scaling to be a valuable tool to tackle this challenge. By constructing large and diverse depth datasets and leveraging powerful foundation models, MDE has become more accessible and robust to real-world use. Foundation models are large neural networks pre-trained on internet-scale data, which allow them to develop a high-level visual understanding that can either be used directly or fine-tuned for a variety of downstream tasks \citep{bommasani2021opportunities}.

\textbf{Ordinal Depth.} One of the earlier works aimed at addressing depth in the wild by leveraging the scale of data is \citep{chen2016single}. Building on this idea, \citet{chen2020oasis} introduced the OASIS dataset, a large-scale dataset specifically designed for depth and normal estimation. It is worth noting, however, both of these works primarily focus on relative (ordinal) depth, which only estimates the depth order instead of providing absolute measurements. While ordinal depth can provide valuable information about the structure of a given scene, its practical use is limited.

\textbf{Affine-invariant Depth.} To overcome the limitations of ordinal depth, several studies have explored affine-invariant depth estimation, which provides depth estimates up to an unknown affine transformation. In other words, the absolute scale and offset of the depth map can vary while the relative depth differences are preserved. For instance, models trained on the MegaDepth dataset \citep{li2018megadepth}, which uses multi-view internet photo collections along with structure-from-motion and multi-view stereo methods to create depth maps, can generalize well to unseen images. Another significant contribution is MiDaS \citep{ranftl2020towards}, which achieves the ability to generalize across a variety of scenes and conditions through training on a mixture of multiple datasets.

\textbf{Metric Depth.} For applications that often require absolute distances, such as robotics \citep{dong2022towards,roussel2019monocular}, augmented reality \citep{kalia2019real}, and autonomous driving \citep{xue2020toward,xiang2022visual}, metric depth estimation is crucial. Metric depth estimation aims to provide absolute depth measurements in real-world units (e.g., meters or centimeters). Inconveniently, zero-shot generalization is particularly challenging due to the metric ambiguities introduced by different camera models. Aside from some works that explicitly incorporate camera intrinsics as an additional input \citep{guizilini2023towards,yin2023metric3d} to directly solve this issue, current state-of-the-art metric depth estimation approaches still rely on fine-tuning powerful foundation models in the respective domain \citep{bhat2023zoedepth,Piccinelli_2024_CVPR,yang2024depth_1,yang2024depth_2}.

\subsection{Uncertainty Quantification}
\textbf{Overview.} A wide variety of UQ methods have been proposed to address the shortcomings of deep neural networks, particularly in terms of reliability and robustness \citep{mackay1992PracticalBayesian,gal2016DropoutBayesian,lakshminarayanan2017SimpleScalable,valdenegro2023sub,van2020uncertainty,liu2020simple,mukhoti2023deep,amini2020deep}. Among them, sampling-based approaches are the most prominent due to their ease of use and effectiveness in providing high-quality uncertainty estimates \citep{mackay1992PracticalBayesian,gal2016DropoutBayesian,lakshminarayanan2017SimpleScalable}. While they usually produce the most accurate uncertainty estimates, the computational cost associated with the necessity of multiple forward passes often makes them unusable for real-world applications that require fast inference times or that are running on resource-constrained devices.

\textbf{Sampling-based Methods.} One of the simplest and most prevalent sampling-based UQ methods is Monte Carlo Dropout (MCD) \citep{gal2016DropoutBayesian}. MCD approximates a Gaussian process by keeping dropout layers active during both training and testing. Originally, dropout layers were solely introduced as a regularization technique to prevent overfitting \citep{srivastava2014Dropout}. With MCD, however, they are also used during test time to turn a deterministic model into a stochastic one to sample from the posterior distribution. The predictive uncertainty of a given model can then easily be estimated by calculating the standard deviation (or variance) of the samples. Another widespread sampling-based UQ method are Deep Ensembles \citep{lakshminarayanan2017SimpleScalable}, which are generally considered the state-of-the-art for UQ across various tasks \citep{ovadia2019DatasetShift,wursthorn2022,gustafsson2020evaluating,landgraf2024evaluation}. They consist of a collection of independently trained models, ideally each initialized with random weights and optimized with random data augmentations to maximize the diversity among the ensemble members \citep{fort2020DeepEnsembles}. The high effectiveness of Deep Ensembles comes at the cost of a very high computational overhead due to the need to train and evaluate multiple models. 

\textbf{Efficiency-oriented Methods.} In addition to these methods, there are several approaches that aim to balance computational efficiency with the quality of uncertainty estimates \citep{valdenegro2023sub,van2020uncertainty,liu2020simple,mukhoti2023deep}. For instance, Sub-Ensembles \citep{valdenegro2023sub} exploit subnetworks within a single model to produce diverse predictions without the need of a full ensemble of models. They offer an effective trade-off between uncertainty quality and computational cost, which can easily be tuned based on the given constraints.

\textbf{UQ in Self-supervised MDE.} While substantial progress has been made in developing effective UQ methods, integrating these techniques with MDE comes with some unique challenges like the limited ground truth data, which is expensive and difficult to acquire, especially at scale. For that reason, a significant body of research has focused on UQ in self-supervised MDE \citep{poggi2020uncertainty,hirose2021variational,nie2021uncertainty,choi2021adaptive,dikov2022variational}. While they all explore different strategies to estimate the uncertainty, some even manage to leverage the uncertainty to enhance model performance \citep{poggi2020uncertainty,nie2021uncertainty,choi2021adaptive}. 

\textbf{UQ in Supervised MDE.} For supervised learning scenarios, one common approach to UQ involves modeling the regression output as a parametric distribution and training the model to estimate its parameters \citep{kendall2017uncertainties,nix1994estimating}. This approach allows the model to not only output the depth estimate but also to measure the corresponding uncertainty, typically represented as the variance of the distribution. Similarly, \citet{yu2021slurp} propose an auxiliary network that exploits the output and the intermediate representations of the main model to estimate the uncertainty. \citet{hornauer2022gradient} introduce a post hoc uncertainty estimation method that relies on gradients extracted with an auxiliary loss function. This technique utilizes Test-Time Augmentation \citep{ayhan2018test} to investigate the correspondence of the depth prediction for an image and its horizontally flipped counterpart. Another innovative approach is proposed by \citet{franchi2022latent}, who address computational efficiency by optimizing a set of latent prototypes. The uncertainty is quantified by examining the position of an input sample in the prototype space. Lastly, \citet{mi2022training} developed a training-free uncertainty estimation approach based on tolerable perturbations during inference and using the variance of multiple outputs as a surrogate for the uncertainty.

\textbf{Research Gap.} Despite significant advancements in UQ for MDE, integrating these techniques with large-scale foundation models remains unexplored. We aim to address this gap by combining multiple UQ methods with the state-of-the-art DepthAnythingV2 foundation model \citep{yang2024depth_2}, enabling pixel-wise uncertainty estimates in addition to metric depth measurements. Our findings will contribute to a more nuanced understanding of the capabilities and limitations of these models, emphasizing the importance of not only striving for higher performance but also making MDE more reliable and trustworthy for real-world use.

\section{DepthAnything Foundation Model}
DepthAnythingV2 \citep{yang2024depth_2} is one the most recent state-of-the-art foundation models for MDE, which can easily be fine-tuned for metric depth estimation, making it the perfect candidate for exploring the combination of various UQ methods with MDE foundation models. Consequently, we want to provide a more detailed overview of this model, which was built upon the framework established by its predecessor, DepthAnythingV1 \citep{yang2024depth_1}.

\textbf{DepthAnythingV1}. \citet{yang2024depth_2} laid the groundwork for creating a versatile foundation model for MDE using the DINOv2 encoder \citep{oquab2023dinov2} for feature extraction with the DPT decoder \citep{ranftl2021vision} for depth regression. The training process of DepthAnythingV1 involves a semi-supervised approach using a student-teacher framework. The teacher model generates pseudo-labels for an extremely large corpus of unlabeled images (approx. 62 million from 8 public datasets), while the student is trained on both the pseudo-labels and a set of 1.5 million labeled images from 6 public datasets. To ensure robustness of the learned representations, they additionally apply strong image perturbations for the student. These include strong color distortion like color jittering and Gaussian blurring and strong spatial distortion through CutMix \citep{yun2019cutmix}. 

To further refine the model's capabilities, they also introduce an auxiliary feature alignment loss. It measures the cosine similarity between the features of the student model and those of a frozen DINOv2 encoder, which is a powerful model for semantic-related tasks like image retrieval and semantic segmentation. This potent addition to the training process helps imbue the DepthAnythingV1 model with high-level semantic understanding to further improve the depth estimation.

\textbf{DepthAnythingV2}. Building on the success of DepthAnythingV1, \citet{yang2024depth_2} quickly introduced several key advancements that enable finer and more robust depth predictions. These improvements are centered around the following three strategies:
\begin{enumerate}
    \item \textbf{Synthetic Data for Label Accuracy:} One of the most significant changes for DepthAnythingV2 is the replacement of all labeled real images with synthetic images. This alteration is motivated by the desire to eliminate label noise and address the lack of detail often ignored in real datasets. In contrast to real images, synthetic data allows for precise depth training and avoids the inconsistencies found in real-world labels.
    \item \textbf{Scaling up the Teacher Model:} To mitigate the drawbacks of synthetic images, such as distribution shifts and restricted scene coverage, the capacity of the teacher model significantly increased. DepthAnythingV2 employs DINOv2-G, the most powerful variant of the DINOv2 encoder \citep{ranftl2021vision}.
    \item \textbf{Leveraging Pseudo-Labeled Real Images:} To bridge the gap between synthetic images and the complexity of real-world scenes, DepthAnythingV2 incorporates large-scale pseudo-labeled real images into its training pipeline. This does not only expand the lacking scene coverage of the synthetic images but also ensures that the model is exposed to a wide variety of real-world scenarios, improving its generalization capabilities.
\end{enumerate}
These advancements make DepthAnythingV2 one of the most powerful, if not the most powerful currently available MDE foundation model. Aside from providing high-quality depth predictions across a diverse range of applications, the model can easily be fine-tuned for metric depth estimation tasks, achieving state-of-the-art results. Given these attributes, DepthAnythingV2 serves as an ideal candidate for exploring foundation model uncertainty in MDE. We aim to assess various UQ methods in combination with this foundation model, enabling pixel-wise uncertainty measures while maintaining high accuracy in the regressing metric depths in real-world applications.

\section{Methodology}
\subsection{Overview}
Our primary research question is straightforward: How can we bridge the gap between ground-breaking results in research and safe deployment in real-world applications that need robust metric depth estimates with corresponding uncertainties.

\begin{figure*}
    \centering
    \includegraphics[width=0.99\linewidth]{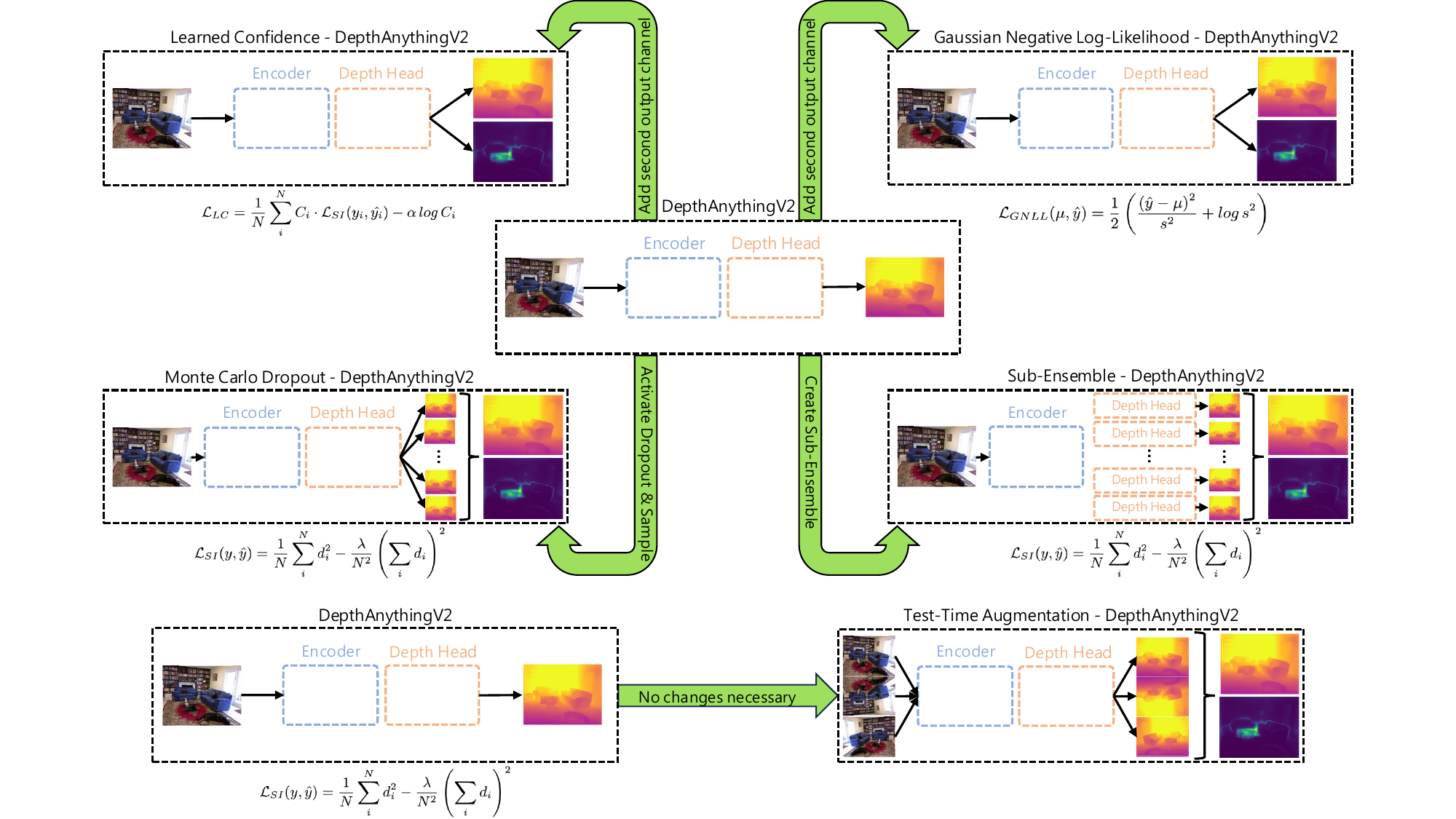}
    \caption{A schematic overview of how to fuse the five different uncertainty quantification approaches with the DepthAnythingV2 \protect\citep{yang2024depth_2} foundation model.}
    \label{fig: methodology}
\end{figure*}

As Figure \ref{fig: methodology} shows, we study five different approaches to not only estimate metric depths but also their corresponding uncertainties: Learned Confidence (LC) \citep{wan2018confnet}, Gaussian Negative Log-Likelihood (GNLL) \citep{nix1994estimating}, Monte Carlo Dropout (MCD) \citep{gal2016DropoutBayesian}, Sub-Ensembles (SE) \citep{valdenegro2023sub}, and Test-Time Augmentation (TTA) \citep{ayhan2018test}

The upper two approaches of Figure \ref{fig: methodology}, \textbf{LC and GNLL}, are fairly simple since they only require adding a second output channel to the already existing depth head. The first output channel outputs the metric depth maps and the second output channel the uncertainty.

The third option, \textbf{MCD}, is equally straightforward since it only requires activating all of the already existing dropout layers in the model while fine-tuning. During inference, these dropout layers are kept active and multiple depth outputs are being sampled. By computing the mean and variance, the final depth map and the corresponding uncertainty are obtained.

The fourth option, \textbf{SE}, is possibly the most complicated and requires significant changes to the architecture. Instead of using just one depth head, a sub-ensemble of randomly initialized depth heads is created. During inference, every depth head predicts slightly different depth samples. Similar to the third option, MCD, the final depth map and uncertainty can be obtained by computing the mean and variance.

Finally, as Figure \ref{fig: methodology} shows, we also examine \textbf{TTA}, which does not require any changes to the fine-tuning process of DepthAnythingV2. Instead, we apply horizontal and vertical flipping during test-time to create two additional inputs to create a total of three unique depth samples. Based on these, we compute the mean and variance to obtain the final depth map and the uncertainty.

\subsection{Learned Confidence}
The general approach of LC was originally proposed by \citet{wan2018confnet} for classification tasks, but has already been adapted before for regression tasks by \citet{wang2024dust3r}. The confidences, which we interpret as uncertainties, can simply be learned in addition to the primary objective function: 

\begin{equation}\label{L_LC}
    \mathcal{L}_{\mathrm{LC}} = \frac{1}{N} \sum_{i}^{N} C_{i} \cdot \mathcal{L}_{\mathrm{SI}}(y_i, \hat{y_i}) - \alpha \, \log \, C_{i} \enspace,
\end{equation}

where $N$ represents the number of pixels having valid ground truth values, $C_i$ denotes the confidence score of the $i$-th valid pixel generated by the model, $y_{i}$ is the depth output of the model, $\hat{y}_{i}$ is the ground truth depth, $\alpha$ is set to 0.2, following \citet{wang2024dust3r}, and $\mathcal{L}_{SI}(y_i, \hat{y_i})$ is the scale-invariant loss introduced by \citet{eigen2014depth}: 

\begin{equation}\label{L_SI}
    \mathcal{L}_{\mathrm{SI}}(y, \hat{y}) = \frac{1}{N} \sum_{i}^{N} d^{2}_{i} - \frac{\lambda}{N^{2}} \left( \sum_{i} d_{i} \right)^{2} \enspace,
\end{equation}

where $d_{i} = \log\,y_i - \log\,\hat{y_i}$. We follow the fine-tuning recommendations of DepthAnythingV2 \citep{yang2024depth_2} and set $\lambda = 0.15$.

\subsection{Gaussian Negative Log-Likelihood Loss}
For depth regression, neural networks are usually only trained to output a predictive mean $\mu$. To also approximate the corresponding variance $s^2$, i.e. the uncertainty, we follow the approach of \citet{nix1994estimating}: By treating the neural network prediction as a sample from a Gaussian distribution, we can minimize the Gaussian Negative Log-Likelihood (GNLL) loss, which can be formulated as:

\begin{equation}
    \mathcal{L}_{\mathrm{GNLL}}(\mu, \hat{y}) = \frac{1}{2} \left( \frac{\left( \hat{y} - \mu \right)^2}{s^2} + \log\,s^2 \right) \enspace .
\end{equation}

Analogous to Equation~\ref{L_LC} for LC, there is no ground truth for the uncertainty, which means that $s^2$ is solely learned implicitly through the optimization of the predictive means $\mu$ based on the ground truth labels $\hat{y}$. 

\subsection{MC Dropout}
Using MCD \citep{gal2016DropoutBayesian} to estimate the predictive mean $\mu$ and the corresponding uncertainty, i.e., the variance $s^2$ or standard deviation $s$, is fairly straightforward. Since the DepthAnythingV2 model already applies dropout layers throughout its architecture, we simply have to activate them not only during training but also during inference to sample from the posterior of the network. 

To compute the predictive mean $\mu$, we take the average of all the samples:

\begin{equation}\label{mean}
    \mu = \frac{1}{T} \sum_{t}^{T} y_t \enspace ,
\end{equation}

where $T$ is the number of samples and $y_t$ is the $t$-th depth prediction of the network. 

For the uncertainty, we calculate the variance:

\begin{equation}\label{var}
    s^2 = \frac{1}{T-1} \sum_{t}^{T} \left( y_t - \mu \right)^2 \enspace .
\end{equation}

Besides that, we follow the fine-tuning recommendations of DepthAnythingV2 \citep{yang2024depth_2}, using the scale-invariant loss $\mathcal{L}_{SI}(y,\hat{y})$ from Equation~\ref{L_SI}  as the objective function.

\subsection{Sub-Ensemble}
Sub-Ensembles (SE) \citep{valdenegro2023sub} enable the approximation of Deep Ensemble. While a Deep Ensemble requires multiple models to be trained and used during inference, the SE only requires a subset of the layers to be multiplied. As shown by Figure \ref{fig: methodology}, we use a shared encoder for multiple randomly initialized depth heads. To maximize the diversity across the depth heads and decrease the training time, we cycle through the heads during training. Per training batch, only one head is optimized, the others are ignored. During inference, however, each depth head predicts a unique sample $y_t$ based on the extracted feature from the encoder. Similar to MCD, we can compute the mean $\mu$ and variance $s^2$ of all the samples (see Equations~\ref{mean} and \ref{var}) to get the desired output, i.e., a final depth map and a corresponding uncertainty.

As for MCD, we follow the fine-tuning recommendations of DepthAnythingV2 \citep{yang2024depth_2}, using the scale-invariant loss $\mathcal{L}_{SI}(y,\hat{y})$ from Equation \ref{L_SI} as the objective function.

\subsection{Test-Time Augmentation}
In contrast to the other four uncertainty quantification approaches, we just fine-tune the DepthAnythingV2 model with the scale-invariant loss $\mathcal{L}_{SI}$ from Equation~\ref{L_SI} for metric depth estimation and apply Test-Time Augmentation (TTA) after training \citep{ayhan2018test}. As shown by Figure \ref{fig: methodology}, we flip the input image vertically as well as horizontally and perform inference with each. As a result, we obtain three unique depth samples $y_t$ that we can use to compute the the mean $\mu$ and variance $s^2$ (see Equations~\ref{mean} and \ref{var}).

\section{Experiments}
\subsection{Experimental Setup}
\textbf{Training}. For all training processes, we follow the default settings of DepthAnythingV2 for metric depth fine-tuning \citep{yang2024depth_2}, using an AdamW optimizer \citep{loshchilov2017decoupled} with a base learning rate of $6\cdot10^{-5}$, a weight decay of 0.01, and a polynomial learning rate scheduler: 

\begin{equation}
lr = lr_\mathrm{base} \cdot (1 - \frac{\mathrm{iteration}}{\mathrm{total\:iterations}})^{0.9}
\enspace, 
\end{equation}

where $lr$ is the current learning rate and $lr_{base}$ is the initial base learning rate. Every model is trained for 25 epochs with an effective batch size of 16, using four NVIDIA A100 GPUs. We do not employ any early stopping techniques and hence only evaluate the final model checkpoints.

\begin{table*}[t!]
\begin{adjustbox}{width=\linewidth}
\setlength\extrarowheight{1mm}
\begin{tabular}{l|l|l|l|l}
                                        & Scene / Application & Resolution (W $\times$ H) & Training Images & Test Images \\ \hline
Cityscapes \citep{cordts2016cityscapes}  & Outdoor           & 2048 $\times$ 1024 & 2975  & 500            \\
NYUv2 \citep{silberman2012indoor}        & Indoor            & 640 $\times$ 480   & 795   & 654            \\
UseGeo \citep{nex2024usegeo}             & Aerial            & 1989 $\times$ 1320 & 551   & 277            \\
HOPE \citep{tyree2022hope}               & Robotics          & 1920 $\times$ 1080 & 49450 (synth.) & 457 (real)        
\end{tabular}
\end{adjustbox}
\caption{Overview of metric depth datasets that we used for evaluation.}
\label{table: datasets}
\end{table*}

\textbf{Datasets.} We conduct our experiments on four highly different datasets, simulating a broad range of real-world applications, as shown by Table \ref{table: datasets}. Cityscapes \citep{cordts2016cityscapes} provides an urban street scene benchmark dataset with high-resolution images. In contrast, NYUv2 \citep{silberman2012indoor} presents indoor scenes with a very low image resolution. UseGeo \citep{nex2024usegeo} covers high-resolution aerial images, which are often neglected in the computer vision community despite their significance in many real-world applications. Finally, the HOPE \citep{tyree2022hope} dataset offers a variety of household objects, originally designed for pose estimation. The main reason why the HOPE dataset is so interesting is that the training dataset is based on almost 50,000 synthetic images, whereas the test dataset consists of just 457 real images. A unique challenge that is often overlooked but fairly common, especially in robotics \citep{hodavn2020bop,sundermeyer2023bop,hodan2024bop}.

\textbf{Data Augmentations.} Regardless of the trained model, we apply random cropping with a crop size of 756 $\times$ 756 pixels on all datasets except NYUv2, which uses 630 $\times$ 476 pixels, and random horizontal flipping with a flip chance of 50\%. For testing purposes, we use the original image resolutions as shown by Table \ref{table: datasets}.

\textbf{Metrics.} For quantitative evaluations of the metric depth estimation, we report all common metrics: root mean squared error (RMSE), absolute relative error (AbsRel), logarithmic root mean squared error (log10), and three threshold-based accuracies ($\delta_1$, $\delta_2$, $\delta_3$). 

To evaluate the uncertainty quality, we exploit the following uncertainty evaluation metrics proposed by \citet{mukhoti2018evaluating}: 

\begin{enumerate}
    \item \textbf{p(accurate|certain):} The probability that the model is accurate on its output given that the uncertainty is below a specified threshold.
    \item \textbf{p(uncertain|inaccurate):} The probability that the uncertainty of the model exceeds a specified threshold given that the prediction is inaccurate.
    \item \textbf{PAvPU:} The combination of metrics 1 and 2. 
\end{enumerate}

Despite the fact that these metrics have originally been proposed for semantic segmentation \citep{mukhoti2018evaluating}, they can also be used to evaluate the depth regression uncertainties \citep{landgraf2024efficient,landgraf2024evaluation}. To determine whether a depth prediction is accurate or inaccurate, we use the strictest threshold-based accuracy $\delta_1$: 

\begin{equation}
    \max\left(\frac{y}{\hat{y}}, \frac{\hat{y}}{y}\right) = \delta_{1} < 1.25
    \enspace.
\end{equation}

To simulate real-world employment, we set the threshold to determine whether a pixel is certain or uncertain to the median uncertainty of a given image \citep{landgraf2024evaluation}.

\textbf{Uncertainty Quantification}. Unless stated otherwise, for MCD and SE, the final depth map and corresponding uncertainty are computed using ten samples or ten depth heads, respectively, in accordance with the findings in \citep{lakshminarayanan2017SimpleScalable,fort2019deep,landgraf2024dudes,landgraf2024uncertainty}.

\subsection{Quantitative Evaluation}\label{sec: quantitative_evaluation}
\begin{table*}[t!]
\begin{center}
\setlength\extrarowheight{1mm}
\begin{tabular}{ll|cc:c:cc}
\multicolumn{2}{c|}{\textbf{NYUv2 (indoor)}} & Trainable Parameters [M] $\downarrow$ & FLOPs [G] $\downarrow$ & Training Time [mm:ss] $\downarrow$ & Inference Time [ms] $\downarrow$ & FPS $\uparrow$ \\ \hline
\multirow{6}{*}{\rotatebox[origin=c]{90}{ViT-S}}
& \multicolumn{1}{|l|}{Baseline} & 24.8 & 66.82 & 00:27 & 12.9 $\pm$ 0.1 & 77.5 \\
& \multicolumn{1}{|l|}{TTA} & 24.8 & 66.82 & 00:27 & 40.0 $\pm$ 0.9 & 25.0 \\ \cdashline{2-6}
& \multicolumn{1}{|l|}{LC} & 24.8 & 66.83 & 00:27 & 12.9 $\pm$ 0.2 & 77.5 \\
& \multicolumn{1}{|l|}{GNLL} & 24.8 & 66.83 & 00:27 & 12.9 $\pm$ 0.2 & 77.5 \\
& \multicolumn{1}{|l|}{SE} & 49.3 & 177.29 & 00:29 & 36.5 $\pm$ 0.5 & 27.4 \\
& \multicolumn{1}{|l|}{MCD} & 24.8 & 66.82 & 00:30 & 128.4 $\pm$ 0.5 & 7.8 \\ \hline
\multirow{6}{*}{\rotatebox[origin=c]{90}{ViT-B}}
& \multicolumn{1}{|l|}{Baseline} & 97.5 & 217.49 & 00:48 & 24.9 $\pm$ 0.5 & 40.2 \\
& \multicolumn{1}{|l|}{TTA} & 97.5 & 217.49 & 00:48 & 75.5 $\pm$ 1.2 & 13.2 \\ \cdashline{2-6}
& \multicolumn{1}{|l|}{LC} & 97.5 & 217.50 & 00:49 & 25.1 $\pm$ 0.5 & 39.8 \\
& \multicolumn{1}{|l|}{GNLL} & 97.5 & 217.50 & 00:49 & 25.0 $\pm$ 0.7 & 40.0 \\
& \multicolumn{1}{|l|}{SE} & 195.5 & 608.09 & 00:52 & 61.9 $\pm$ 0.8 & 16.2 \\
& \multicolumn{1}{|l|}{MCD} & 97.5 & 217.49 & 00:54 & 248.4 $\pm$ 2.7 & 4.0 \\ \hline
\multirow{6}{*}{\rotatebox[origin=c]{90}{ViT-L}}
& \multicolumn{1}{|l|}{Baseline} & 335.3 & 741.40 & 02:11 & 57.1 $\pm$ 3.5 & 17.5 \\
& \multicolumn{1}{|l|}{TTA} & 335.3 & 741.40 & 02:11 & 172.6 $\pm$ 11.6 & 5.8 \\ \cdashline{2-6}
& \multicolumn{1}{|l|}{LC} & 335.3 & 741.41 & 02:14 & 57.0 $\pm$ 3.9 & 17.5 \\
& \multicolumn{1}{|l|}{GNLL} & 335.3 & 741.41 & 02:13 & 57.1 $\pm$ 3.3 & 17.5 \\
& \multicolumn{1}{|l|}{SE} & 613.8 & 2204.07 & 02:17 & 131.6 $\pm$ 5.2 & 7.6 \\
& \multicolumn{1}{|l|}{MCD} & 335.3 & 741.40 & 02:26 & 569.8 $\pm$ 24.3 & 1.8 \\ \hline
\end{tabular}
\end{center}
\caption{Efficiency comparison between the five chosen uncertainty quantification methods: Test-Time Augmentation (TTA), Learned Confidence (LC), Gaussian Negative Log-Likelihood (GNLL), Sub-Ensemble (SE), and MC Dropout (MCD) for three different encoder sizes: ViT-S, ViT-B, ViT-L on the NYUv2 \protect\citep{silberman2012indoor} dataset. We compare the number of trainable parameters, FLOPs, training time per epoch on a single A100 GPU, inference time per image, and the respective FPS. The mean inference time and corresponding standard deviation are based on 1000 forward passes.}
\label{table: inference}
\end{table*}

\textbf{Efficiency.} 
As shown by Table \ref{table: inference}, training times are comparable to the baseline for all methods, with LC and GNLL matching it exactly in both training and inference time. SE and MCD increase training times by around 5\% - 10\%. While SE nearly triples inference time and roughly doubles the trainable parameters, MCD requires roughly 10 times the inference time due to the costly sampling process that roughly scales linearly with the amount of samples. TTA also triples inference time as it requires three forward passes. Overall, LC and GNLL are the most efficient approaches as they do not require any additional computational overhead compared to the baseline. These experiments were conducted on the NYUv2 dataset only, so exact numbers may vary across datasets, but the general findings should be representative. 

\begin{table*}[t!]
\begin{center}
\begin{adjustbox}{width=\linewidth}
\setlength\extrarowheight{1mm}
\begin{tabular}{ll|cccccc:ccc}
\multicolumn{2}{c|}{\textbf{NYUv2 (indoor)}} & RMSE $\downarrow$ & AbsRel $\downarrow$ & log10 $\downarrow$ & $\delta_1$ $\uparrow$ & $\delta_2$ $\uparrow$ & $\delta_3$ $\uparrow$ & p(acc|cer) $\uparrow$ & p(unc|ina) $\uparrow$ & PAvPU $\uparrow$ \\ \hline
\multirow{6}{*}{\rotatebox[origin=c]{90}{ViT-S}}
& \multicolumn{1}{|l|}{Baseline} & \textbf{0.340} & 0.093 & 0.039 & 0.928 & 0.988 & 0.997 & - & - & - \\
& \multicolumn{1}{|l|}{TTA} & 0.399 & 0.111 & 0.049 & 0.881 & 0.984 & 0.997 & 0.903 & 0.602 & 0.522 \\ \cdashline{2-11}
& \multicolumn{1}{|l|}{LC} & 0.343 & 0.090 & 0.039 & 0.930 & 0.988 & 0.997 & 0.920 & 0.369 & 0.490 \\
& \multicolumn{1}{|l|}{GNLL} & 0.342 & 0.094 & 0.040 & 0.924 & 0.987 & 0.997 & \textbf{0.953} & \textbf{0.846} & 0.529 \\
& \multicolumn{1}{|l|}{SE} & \textbf{0.340} & 0.092 & 0.039 & 0.926 & 0.988 & 0.997 & 0.937 & 0.704 & 0.511 \\
& \multicolumn{1}{|l|}{MCD  (10\%)} & 0.422 & 0.121 & 0.050 & 0.867 & 0.973 & 0.992 & 0.900 & 0.692 & \textbf{0.533} \\ \hline
\multirow{6}{*}{\rotatebox[origin=c]{90}{ViT-B}}
& \multicolumn{1}{|l|}{Baseline} & 0.307 & 0.080 & 0.034 & 0.948 & 0.991 & 0.998 & - & - & - \\
& \multicolumn{1}{|l|}{TTA} & 0.359 & 0.099 & 0.435 & 0.910 & 0.988 & 0.998 & 0.924 & 0.599 & 0.514 \\ \cdashline{2-11}
& \multicolumn{1}{|l|}{LC} & 0.314 & 0.085 & 0.036 & 0.943 & 0.991 & 0.998 & 0.926 & 0.292 & 0.483 \\
& \multicolumn{1}{|l|}{GNLL} & \textbf{0.305} & 0.079 & 0.034 & 0.949 & 0.991 & 0.998 & \textbf{0.966} & \textbf{0.826} & 0.517 \\
& \multicolumn{1}{|l|}{SE} & 0.309 & 0.080 & 0.034 & 0.947 & 0.991 & 0.998 & 0.955 & 0.698 & 0.507\\
& \multicolumn{1}{|l|}{MCD  (10\%)} & 0.339 & 0.091 & 0.039 & 0.925 & 0.986 & 0.997 & 0.952 & 0.774 & \textbf{0.527} \\ \hline
\multirow{6}{*}{\rotatebox[origin=c]{90}{ViT-L}}
& \multicolumn{1}{|l|}{Baseline} & \textbf{0.270} & 0.068 & 0.030 & 0.964 & 0.993 & 0.998 & - & - & - \\ 
& \multicolumn{1}{|l|}{TTA} & 0.324 & 0.087 & 0.039 & 0.938 & 0.992 & 0.998 & 0.944 & 0.598 & 0.507 \\ \cdashline{2-11}
& \multicolumn{1}{|l|}{LC} & 0.275 & 0.069 & 0.030 & 0.963 & 0.993 & 0.998 & 0.946 & 0.268 & 0.483\\
& \multicolumn{1}{|l|}{GNLL} & 0.285 & 0.072 & 0.031 & 0.959 & 0.992 & 0.998 & \textbf{0.980} & \textbf{0.912} & 0.521 \\
& \multicolumn{1}{|l|}{SE} & 0.280 & 0.070 & 0.030 & 0.961 & 0.993 & 0.998 & 0.969 & 0.734 & 0.508 \\
& \multicolumn{1}{|l|}{MCD  (10\%)} & 0.339 & 0.091 & 0.039 & 0.925 & 0.986 & 0.997 & 0.967 & 0.798 & \textbf{0.522} \\ 
\end{tabular}
\end{adjustbox}
\end{center}
\caption{Quantitative comparison on the NYUv2 \citep{silberman2012indoor} dataset between the five chosen uncertainty quantification methods: Test-Time Augmentation (TTA), Learned Confidence (LC), Gaussian Negative Log-Likelihood (GNLL), Sub-Ensemble (SE), and MC Dropout (MCD) for three different encoder sizes: ViT-S, ViT-B, ViT-L. Best results for RMSE and the three uncertainty metrics are marked in \textbf{bold} for each encoder.}
\label{table: uq nyuv2}
\end{table*}

\textbf{NYUv2.}
Table \ref{table: uq nyuv2} shows a quantitative comparison for the NYUv2 dataset \citep{silberman2012indoor}. LC, GNLL, and SE maintain depth quality similar to the baseline, with SE and GNLL even surpassing it for the ViT-S and ViT-B encoders, respectively. TTA and MCD exhibit a slight degradation but remain competitive. 

Regarding the uncertainty quality, GNLL emerges as the top-performing method, consistently outperforms all others in terms of p(acc|cer) and p(unc|ina) across all three encoder sizes, achieving impressive values of up to 98.0\% and 91.2\%, respectively. For PAvPU, both GNLL and MCD deliver the best results. While all other methods are somewhat competitive with each other, LC clearly falls behind with regard to p(unc|ina), achieving only 26.8\% for ViT-L. 

\begin{table*}[t!]
\begin{center}
\begin{adjustbox}{width=\linewidth}
\setlength\extrarowheight{1mm}
\begin{tabular}{ll|cccccc:ccc}
\multicolumn{2}{c|}{\textbf{Cityscapes (outdoor)}} & RMSE $\downarrow$ & AbsRel $\downarrow$ & log10 $\downarrow$ & $\delta_1$ $\uparrow$ & $\delta_2$ $\uparrow$ & $\delta_3$ $\uparrow$ & p(acc|cer) $\uparrow$ & p(unc|ina) $\uparrow$ & PAvPU $\uparrow$ \\ \hline
\multirow{6}{*}{\rotatebox[origin=c]{90}{ViT-S}}
& \multicolumn{1}{|l|}{Baseline} & 7.138 & 0.219 & 0.084 & 0.704 & 0.964 & 0.991 & - & - & - \\ 
& \multicolumn{1}{|l|}{TTA} & \textbf{6.474} & 0.223 & 0.089 & 0.576 & 0.965 & 0.991 & 0.335 & 0.421 & 0.410 \\ \cdashline{2-11}
& \multicolumn{1}{|l|}{LC} & 7.492 & 0.206 & 0.078 & 0.733 & 0.958 & 0.991 & 0.658 & 0.619 & 0.611 \\
& \multicolumn{1}{|l|}{GNLL} & 7.739 & 0.236 & 0.089 & 0.669 & 0.949 & 0.987 & \textbf{0.692} & \textbf{0.706} & \textbf{0.695} \\
& \multicolumn{1}{|l|}{SE} & 7.714 & 0.234 & 0.088 & 0.681 & 0.954 & 0.990 & 0.658 & 0.654 & 0.649 \\
& \multicolumn{1}{|l|}{MCD (10\%)} & 8.527 & 0.307 & 0.113 & 0.417 & 0.920 & 0.984 & 0.328 & 0.525 & 0.516 \\ \hline
\multirow{6}{*}{\rotatebox[origin=c]{90}{ViT-B}}
& \multicolumn{1}{|l|}{Baseline} & 6.884 & 0.252 & 0.095 & 0.599 & 0.965 & 0.992 & - & - & - \\ 
& \multicolumn{1}{|l|}{TTA} & \textbf{5.757} & 0.247 & 0.095 & 0.526 & 0.967 & 0.993 & 0.288 & 0.420 & 0.396 \\ \cdashline{2-11}
& \multicolumn{1}{|l|}{LC} & 7.711 & 0.285 & 0.107 & 0.434 & 0.958 & 0.991 & 0.329 & 0.506 & 0.502 \\
& \multicolumn{1}{|l|}{GNLL} & 7.092 & 0.244 & 0.094 & 0.613 & 0.966 & 0.991 & \textbf{0.594} & \textbf{0.635} & \textbf{0.629} \\
& \multicolumn{1}{|l|}{SE} & 7.824 & 0.271 & 0.102 & 0.534 & 0.957 & 0.991 & 0.490 & 0.587 & 0.588 \\
& \multicolumn{1}{|l|}{MCD (10\%)} & 8.268 & 0.288 & 0.107 & 0.488 & 0.939 & 0.989 & 0.449 & 0.579 & 0.583 \\ \hline
\multirow{6}{*}{\rotatebox[origin=c]{90}{ViT-L}}
& \multicolumn{1}{|l|}{Baseline} & 6.655 & 0.256 & 0.097 & 0.558 & 0.972 & 0.993 & - & - & - \\ 
& \multicolumn{1}{|l|}{TTA} & \textbf{5.298} & 0.227 & 0.088 & 0.608 & 0.979 & 0.994 & 0.371 & 0.431 & 0.416 \\ \cdashline{2-11}
& \multicolumn{1}{|l|}{LC} & 7.392 & 0.280 & 0.105 & 0.416 & 0.969 & 0.993 & 0.292 & 0.488 & 0.488 \\
& \multicolumn{1}{|l|}{GNLL} & 6.562 & 0.234 & 0.092 & 0.628 & 0.970 & 0.990 & \textbf{0.581} & \textbf{0.620} & \textbf{0.607} \\
& \multicolumn{1}{|l|}{SE} & 7.522 & 0.272 & 0.103 & 0.500 & 0.966 & 0.993 & 0.446 & 0.568 & 0.570 \\
& \multicolumn{1}{|l|}{MCD (10\%)} & 8.268 & 0.288 & 0.107 & 0488 & 0.939 & 0.989 & 0.480 & 0.584 & 0.584 \\ 
\end{tabular}
\end{adjustbox}
\end{center}
\caption{Quantitative comparison on the Cityscapes \citep{cordts2016cityscapes} dataset between the five chosen uncertainty quantification methods: Test-Time Augmentation (TTA), Learned Confidence (LC), Gaussian Negative Log-Likelihood (GNLL), Sub-Ensemble (SE), and MC Dropout (MCD) for three different encoder sizes: ViT-S, ViT-B, ViT-L. Best results for RMSE and the three uncertainty metrics are marked in \textbf{bold} for each encoder.}
\label{table: uq cityscapes}
\end{table*}

\textbf{Cityscapes.}
For the Cityscapes dataset \citep{cordts2016cityscapes}, as shown by Table \ref{table: uq cityscapes}, TTA stands out by significantly outperforming the baseline across all three encoder sizes in terms of depth quality. In contrast, LC, GNLL, and SE exhibit noticeable degradation, while MCD performs the worst. 

For uncertainty quality, GNLL is the standout performer, decisively surpassing all other methods for all three metrics and encoder sizes. It achieves remarkable values of 69.2\% for p(acc|cer), 70.6\% for p(unc|ina), and 69.5\% for PAvPU, setting a benchmark for reliability in this dataset. The remaining methods deliver less consistent results, with TTA generally performing the worst, showing values as low as 28.8\% for p(acc|cer), 42.0\% for p(unc|ina), and 39.6\% for PAvPU. Interestingly, GNLL remains resilient despite the observed uncertainty degradation with increasing encoder size for most other methods.

\begin{table*}[t!]
\begin{center}
\begin{adjustbox}{width=\linewidth}
\setlength\extrarowheight{1mm}
\begin{tabular}{ll|cccccc:ccc}
\multicolumn{2}{c|}{\textbf{UseGeo (aerial)}} & RMSE $\downarrow$ & AbsRel $\downarrow$ & log10 $\downarrow$ & $\delta_1$ $\uparrow$ & $\delta_2$ $\uparrow$ & $\delta_3$ $\uparrow$ & p(acc|cer) $\uparrow$ & p(unc|ina) $\uparrow$ & PAvPU $\uparrow$ \\ \hline

\multirow{6}{*}{\rotatebox[origin=c]{90}{ViT-S}}
& \multicolumn{1}{|l|}{Baseline} & 7.366 & 0.077 & 0.032 & 0.973 & 0.994 & 0.999 & - & - & - \\ 
& \multicolumn{1}{|l|}{TTA} & 7.078 & 0.074 & 0.031 & 0.977 & 0.995 & 0.999 & 0.976 & 0.647 & 0.500 \\ \cdashline{2-11}
& \multicolumn{1}{|l|}{LC} & 8.213 & 0.086 & 0.036 & 0.958 & 0.995 & 1.000 & 0.963 & 0.481 & \textbf{0.506} \\
& \multicolumn{1}{|l|}{GNLL} & 7.467 & 0.076 & 0.032 & 0.979 & 0.993 & 0.998 & 0.976 & 0.237 & 0.500 \\
& \multicolumn{1}{|l|}{SE} & 7.259 & 0.076 & 0.032 & 0.976 & 0.994 & 0.999 & 0.971 & 0.467 & 0.495 \\
& \multicolumn{1}{|l|}{MCD  (10\%)} & \textbf{6.682} & 0.068 & 0.029 & 0.982 & 0.994 & 0.998 & \textbf{0.982} & \textbf{0.672} & 0.501 \\ \hline

\multirow{6}{*}{\rotatebox[origin=c]{90}{ViT-B}}
& \multicolumn{1}{|l|}{Baseline} & 6.386 & 0.067 & 0.028 & 0.981 & 0.995 & 0.999 & - & - & - \\ 
& \multicolumn{1}{|l|}{TTA} & \textbf{6.060} & 0.631 & 0.027 & 0.985 & 0.995 & 0.999 & \textbf{0.984} & 0.609 & 0.499 \\ \cdashline{2-11}
& \multicolumn{1}{|l|}{LC} & 6.612 & 0.068 & 0.029 & 0.975 & 0.995 & 1.000 & 0.967 & 0.454 & 0.492 \\
& \multicolumn{1}{|l|}{GNLL} & 7.810 & 0.080 & 0.035 & 0.972 & 0.990 & 0.997 & 0.971 & 0.293 & 0.499 \\
& \multicolumn{1}{|l|}{SE} & 6.491 & 0.068 & 0.028 & 0.980 & 0.994 & 0.999 & 0.981 & 0.559 & 0.502 \\
& \multicolumn{1}{|l|}{MCD  (10\%)} & 6.641 & 0.070 & 0.029 & 0.978 & 0.994 & 0.999 & 0.982 & \textbf{0.657} & \textbf{0.504} \\ \hline

\multirow{6}{*}{\rotatebox[origin=c]{90}{ViT-L}}
& \multicolumn{1}{|l|}{Baseline} & 6.173 & 0.065 & 0.027 & 0.981 & 0.995 & 0.999 & - & - & - \\ 
& \multicolumn{1}{|l|}{TTA} & 5.898 & 0.063 & 0.026 & 0.982 & 0.995 & 0.999 & 0.980 & 0.596 & 0.499 \\ \cdashline{2-11}
& \multicolumn{1}{|l|}{LC} & \textbf{5.406} & 0.056 & 0.024 & 0.986 & 0.995 & 1.000 & \textbf{0.985} & 0.477 & 0.500 \\
& \multicolumn{1}{|l|}{GNLL} & 7.082 & 0.073 & 0.031 & 0.980 & 0.991 & 0.998 & 0.980 & 0.294 & 0.498 \\
& \multicolumn{1}{|l|}{SE} & 6.260 & 0.067 & 0.028 & 0.980 & 0.995 & 1.000 & 0.977 & 0.577 & 0.497 \\
& \multicolumn{1}{|l|}{MCD  (10\%)} & 6.697 & 0.071 & 0.029 & 0.981 & 0.995 & 0.999 & 0.982 & \textbf{0.669} & \textbf{0.501} \\ 

\end{tabular}
\end{adjustbox}
\end{center}
\caption{Quantitative comparison on the UseGeo \citep{nex2024usegeo} dataset between the five chosen uncertainty quantification methods: Test-Time Augmentation (TTA), Learned Confidence (LC), Gaussian Negative Log-Likelihood (GNLL), Sub-Ensemble (SE), and MC Dropout (MCD) for three different encoder sizes: ViT-S, ViT-B, ViT-L. Best results for RMSE and the three uncertainty metrics are marked in \textbf{bold} for each encoder.}
\label{table: uq usegeo}
\end{table*}

\textbf{UseGeo.}
For the UseGeo dataset \citep{nex2024usegeo}, as presented by Table \ref{table: uq usegeo}, the depth quality results are inconsistent, with methods sometimes surpassing and at other times falling short of the baseline. TTA is the only approach that consistently outperforms the baseline across all three encoder sizes.

In terms of uncertainty quality, all methods deliver near-perfect results for p(acc|cer) of at least 96.3\%. For p(unc|ina) and PAvPU, MCD generally performs best with values of up 67.2\% and 50.4\%, respectively. In contrast to the other datasets, GNLL significantly lags behind the other methods on UseGeo. This is likely due to the large depth values in UseGeo, which led to much higher absolute GNLL loss values. The GNLL loss incorporates a logarithmic term that penalizes high uncertainty estimates, and with large depth values, the uncertainties—and hence the loss—are naturally magnified. Notably, no hyperparameter adjustments were made to address this, ensuring comparability but potentially hindering GNLL’s optimization in this particular case.

\begin{table*}[t!]
\begin{center}
\begin{adjustbox}{width=\linewidth}
\setlength\extrarowheight{1mm}
\begin{tabular}{ll|cccccc:ccc}
\multicolumn{2}{c|}{\textbf{HOPE (robotics)}} & RMSE $\downarrow$ & AbsRel $\downarrow$ & log10 $\downarrow$ & $\delta_1$ $\uparrow$ & $\delta_2$ $\uparrow$ & $\delta_3$ $\uparrow$ & p(acc|cer) $\uparrow$ & p(unc|ina) $\uparrow$ & PAvPU $\uparrow$ \\ \hline

\multirow{6}{*}{\rotatebox[origin=c]{90}{ViT-S}}
& \multicolumn{1}{|l|}{Baseline} & 0.263 & 0.265 & 0.115 & 0.537 & 0.821 & 0.942 & - & - & - \\ 
& \multicolumn{1}{|l|}{TTA} & 0.264 & 0.262 & 0.114 & 0.539 & 0.818 & 0.943 & \textbf{0.421} & 0.564 & 0.552 \\ \cdashline{2-11}
& \multicolumn{1}{|l|}{LC} & \textbf{0.259} & 0.262 & 0.114 & 0.537 & 0.822 & 0945 & 0.339 & 0.476 & 0.474 \\
& \multicolumn{1}{|l|}{GNLL} & 0.277 & 0.287 & 0.124 & 0.492 & 0.795 & 0.929 & 0.404 & \textbf{0.575} & \textbf{0.567} \\
& \multicolumn{1}{|l|}{SE} & 0.262 & 0.263 & 0.117 & 0.544 & 0.809 & 0.933 & 0.393 & 0.528 & 0.522 \\
& \multicolumn{1}{|l|}{MCD  (10\%)} & 0.274 & 0.286 & 0.118 & 0.514 & 0.816 & 0.937 & 0.398 & 0.562 & 0.553 \\ \hline

\multirow{6}{*}{\rotatebox[origin=c]{90}{ViT-B}}
& \multicolumn{1}{|l|}{Baseline} & 0.222 & 0.232 & 0.094 & 0.616 & 0.892 & 0.969 & - & - & - \\ 
& \multicolumn{1}{|l|}{TTA} & 0.221 & 0.230 & 0.093 & 0.621 & 0.891 & 0.968 & 0.462 & 0.573 & 0.558 \\ \cdashline{2-11}
& \multicolumn{1}{|l|}{LC} & 0.224 & 0.227 & 0.095 & 0.616 & 0.883 & 0.966 & 0.330 & 0.419 & 0.427 \\
& \multicolumn{1}{|l|}{GNLL} & 0.223 & 0.225 & 0.094 & 0.619 & 0.892 & 0.972 & \textbf{0.497} & \textbf{0.622} & \textbf{0.596} \\
& \multicolumn{1}{|l|}{SE} & \textbf{0.218} & 0.230 & 0.094 & 0.625 & 0.889 & 0.967 & 0.448 & 0.546 & 0.543 \\
& \multicolumn{1}{|l|}{MCD  (10\%)} & 0.245 & 0.269 & 0.106 & 0.560 & 0.851 & 0.955 & 0.405 & 0.555 & 0.543 \\ \hline

\multirow{6}{*}{\rotatebox[origin=c]{90}{ViT-L}}
& \multicolumn{1}{|l|}{Baseline} & 0.223 & 0.238 & 0.096 & 0.588 & 0.906 & 0.980 & - & - & - \\ 
& \multicolumn{1}{|l|}{TTA} & 0.217 & 0.232 & 0.094 & 0.599 & 0.904 & 0.980 & 0.436 & 0.576 & 0.557 \\ \cdashline{2-11}
& \multicolumn{1}{|l|}{LC} & \textbf{0.215} & 0.226 & 0.092 & 0.604 & 0.911 & 0.980 & 0.325 & 0.426 & 0.441 \\
& \multicolumn{1}{|l|}{GNLL} & 0.229 & 0.244 & 0.099 & 0.588 & 0.888 & 0.973 & \textbf{0.460} & \textbf{0.608} & \textbf{0.586} \\
& \multicolumn{1}{|l|}{SE} & 0.235 & 0.252 & 0.098 & 0.594 & 0.893 & 0.974 & 0.442 & 0.574 & 0.569 \\
& \multicolumn{1}{|l|}{MCD  (10\%)} & 0.249 & 0.272 & 0.107 & 0.557 & 0.859 & 0.966 & 0.416 & 0.580 & 0.563 \\ 

\end{tabular}
\end{adjustbox}
\end{center}
\caption{Quantitative comparison on the HOPE \citep{tyree2022hope} dataset between the five chosen uncertainty quantification methods: Test-Time Augmentation (TTA), Learned Confidence (LC), Gaussian Negative Log-Likelihood (GNLL), Sub-Ensemble (SE), and MC Dropout (MCD) for three different encoder sizes: ViT-S, ViT-B, ViT-L. Best results for RMSE and the three uncertainty metrics are marked in \textbf{bold} for each encoder.}
\label{table: uq hope}
\end{table*}

\textbf{HOPE.}
Table \ref{table: uq hope} shows the final quantitative comparison for the HOPE dataset \citep{tyree2022hope}. In terms of depth quality, there are only marginal differences between the baseline and all uncertainty approaches, with RMSE values ranging between 0.215 to 0.277.

Regarding the uncertainty quality, GNLL once again asserts itself as the best option for all three metrics across all three encoder sizes, with only one minor exception. GNLL achieves results of up to 49.7\% for p(acc|cer), 62.2\% for p(unc|ina), and 59.6\% for PAvPU. Mirroring its strong performance on NYUv2 (cf. Table \ref{table: uq nyuv2}), GNLL’s dominance is evident, while the other methods remain fairly competitive with each other. LC, however, lags significantly behind across all three uncertainty quality metrics, underscoring GNLL’s reliability and consistency.

\subsection{Qualitative Evaluation}
\begin{figure*}[ht!]
    \centering
    \begin{minipage}[c]{0.025\textwidth}
        \centering
        \adjustbox{valign=c}{\rotatebox[origin=c]{90}{GNLL (ViT-S)}}
    \end{minipage}%
    \begin{minipage}[c]{0.19\textwidth}
        \includegraphics[width=\textwidth]{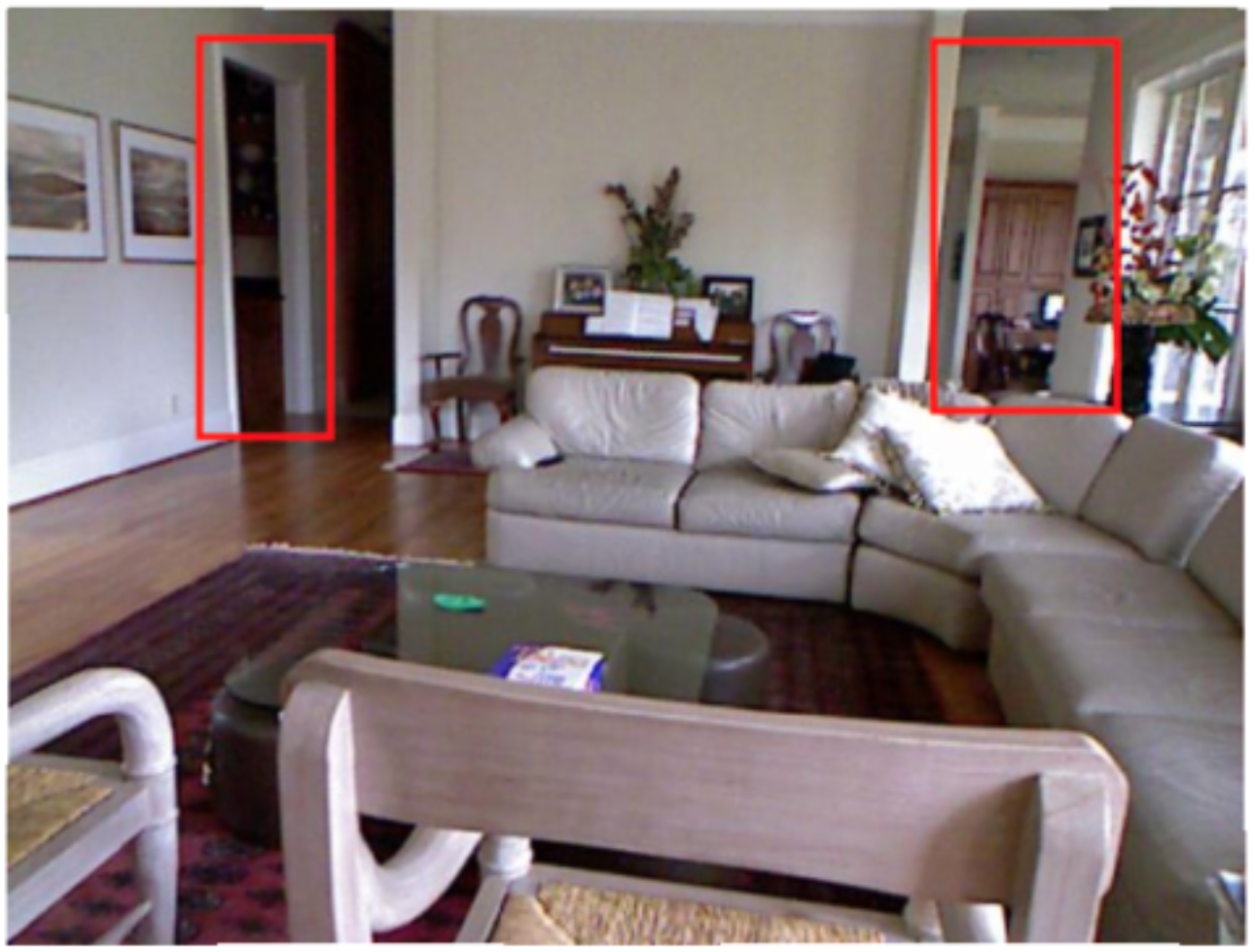}
    \end{minipage}%
    \begin{minipage}[c]{0.19\textwidth}
        \includegraphics[width=\textwidth]{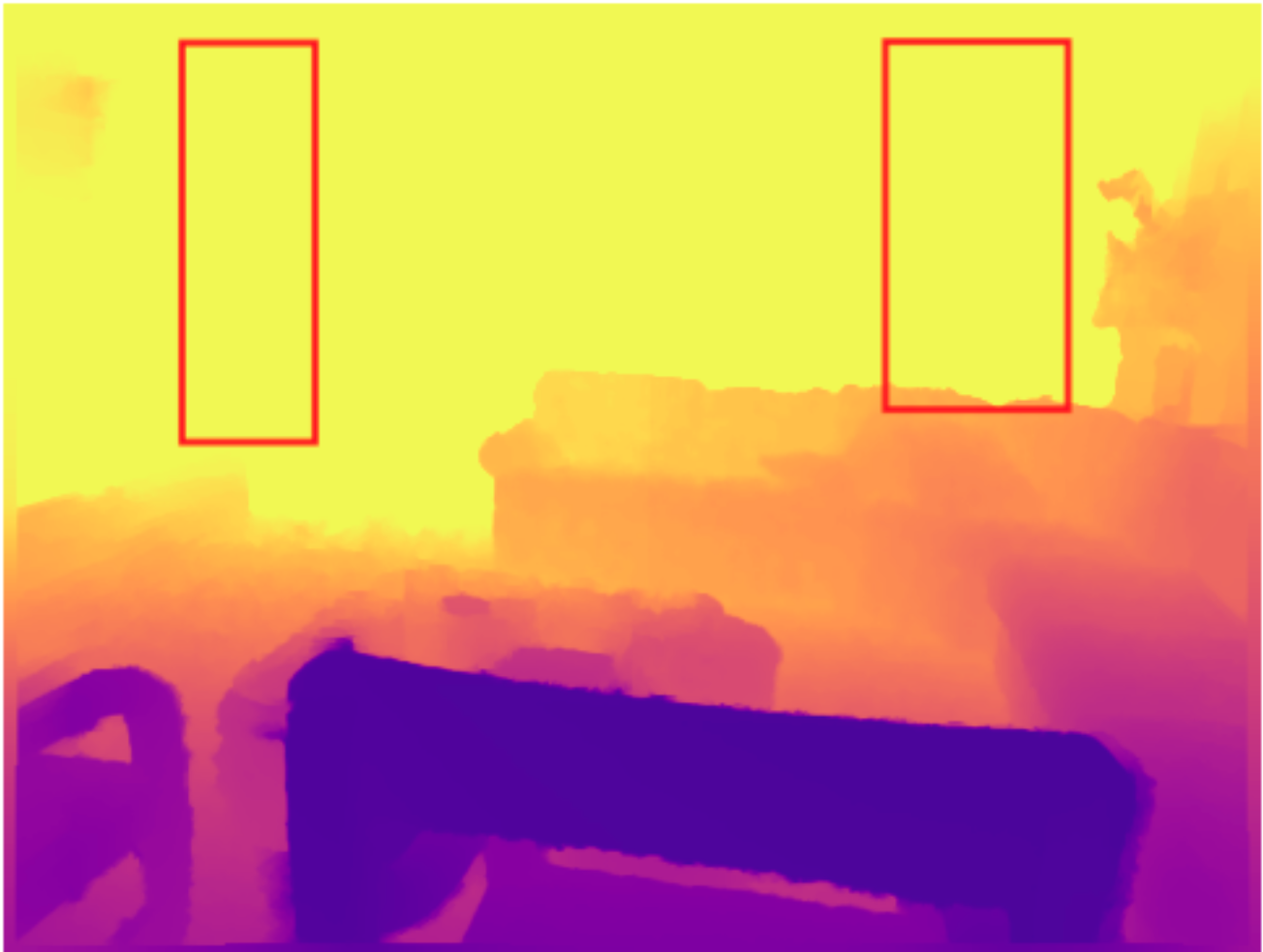}
    \end{minipage}%
    \begin{minipage}[c]{0.19\textwidth}
        \includegraphics[width=\textwidth]{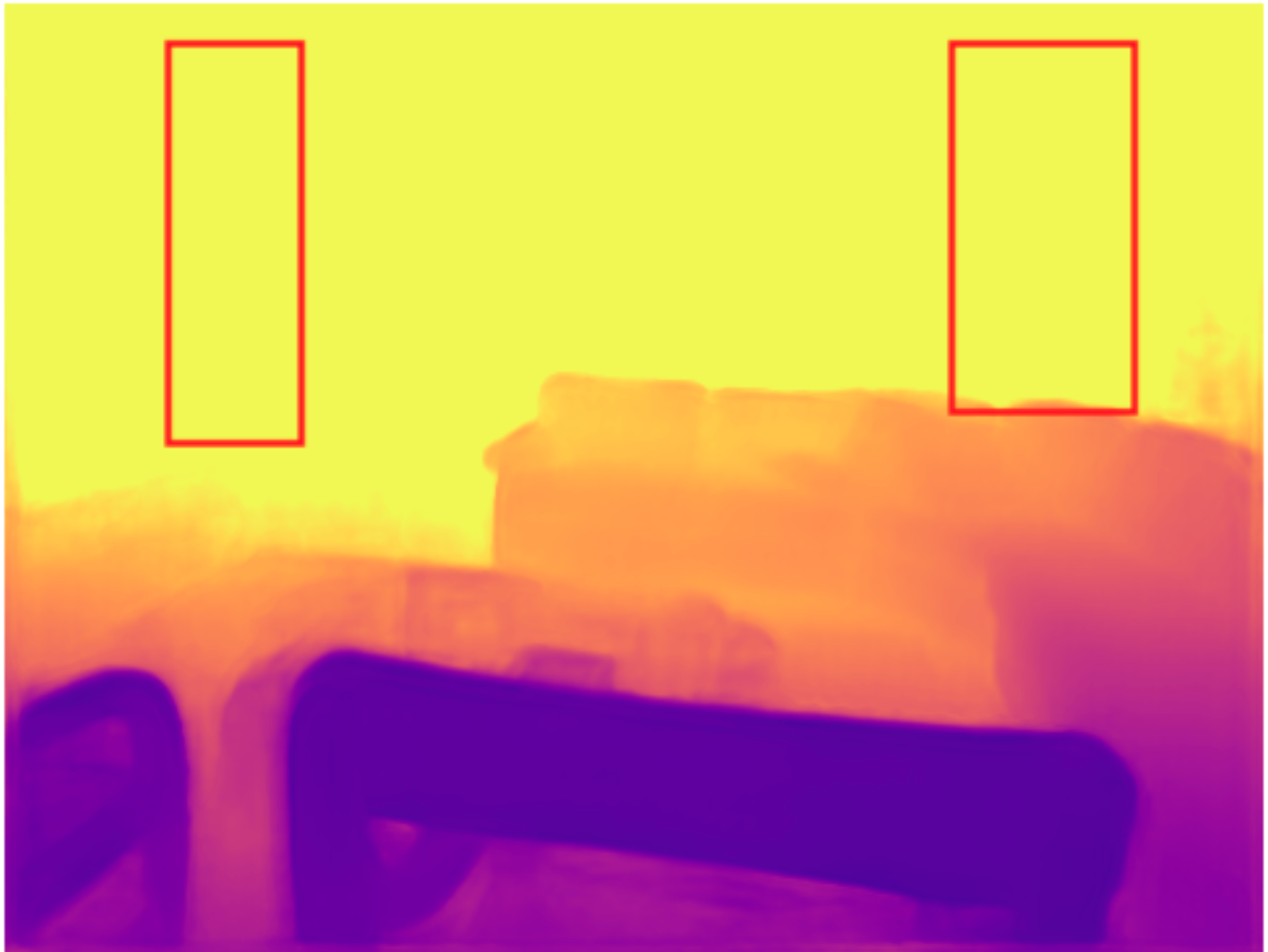}
    \end{minipage}%
    \begin{minipage}[c]{0.19\textwidth}
        \includegraphics[width=\textwidth]{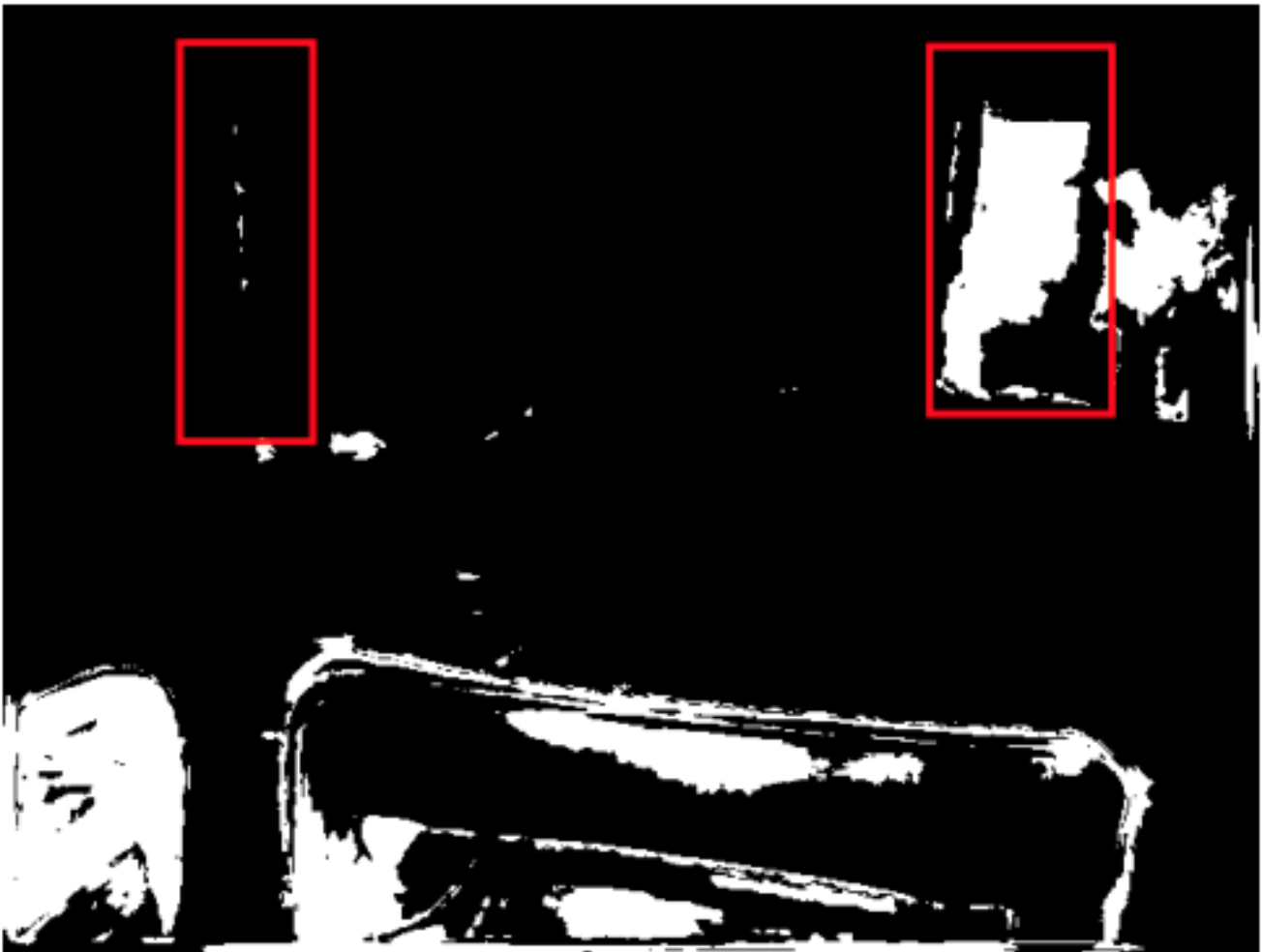}
    \end{minipage}%
    \begin{minipage}[c]{0.19\textwidth}
        \includegraphics[width=\textwidth]{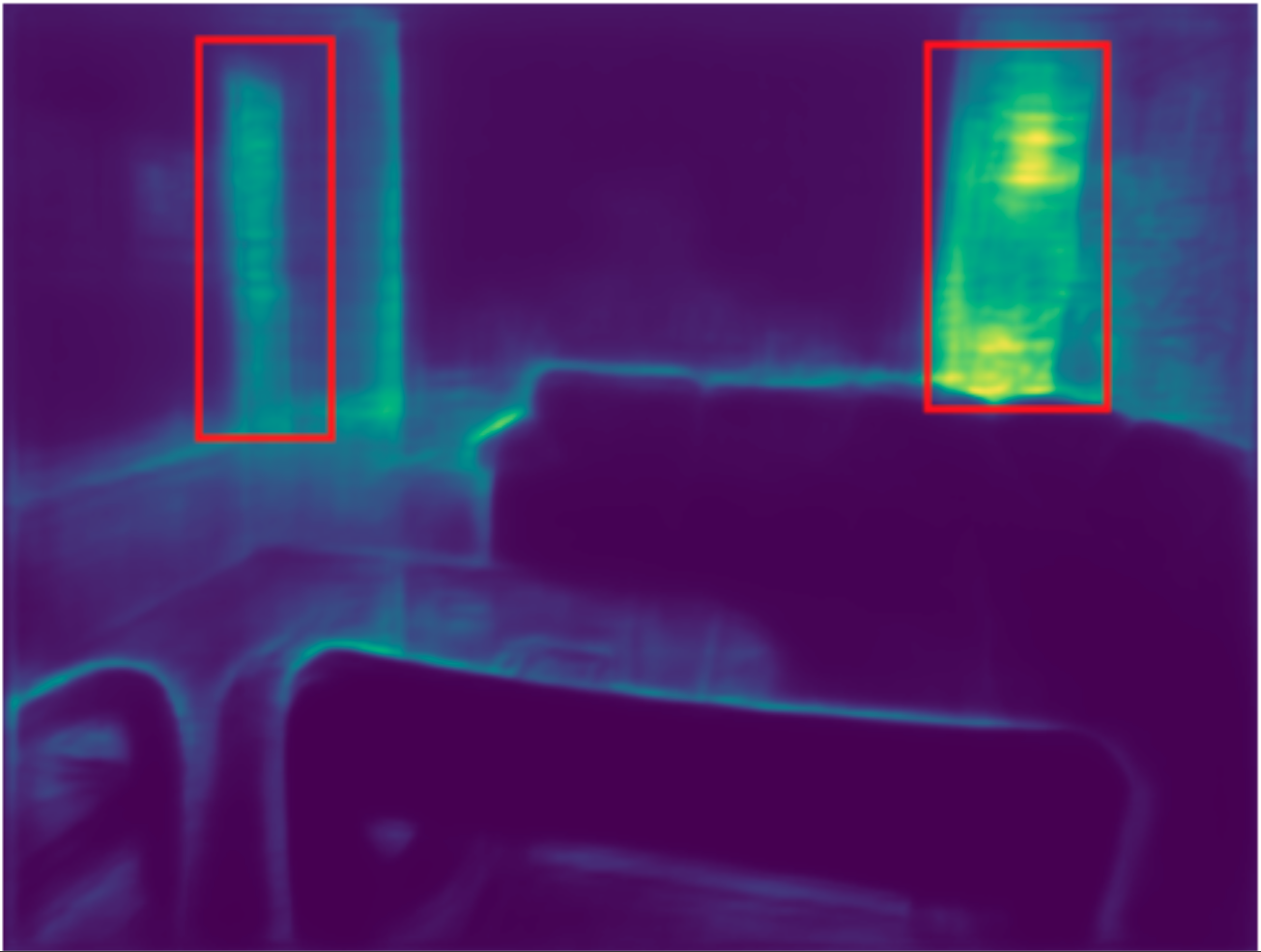}
    \end{minipage}

    \begin{minipage}[c]{0.025\textwidth}
        \centering
        \adjustbox{valign=c}{\rotatebox[origin=c]{90}{TTA (ViT-S)}}
    \end{minipage}%
    \begin{minipage}[c]{0.19\textwidth}
        \includegraphics[width=\textwidth]{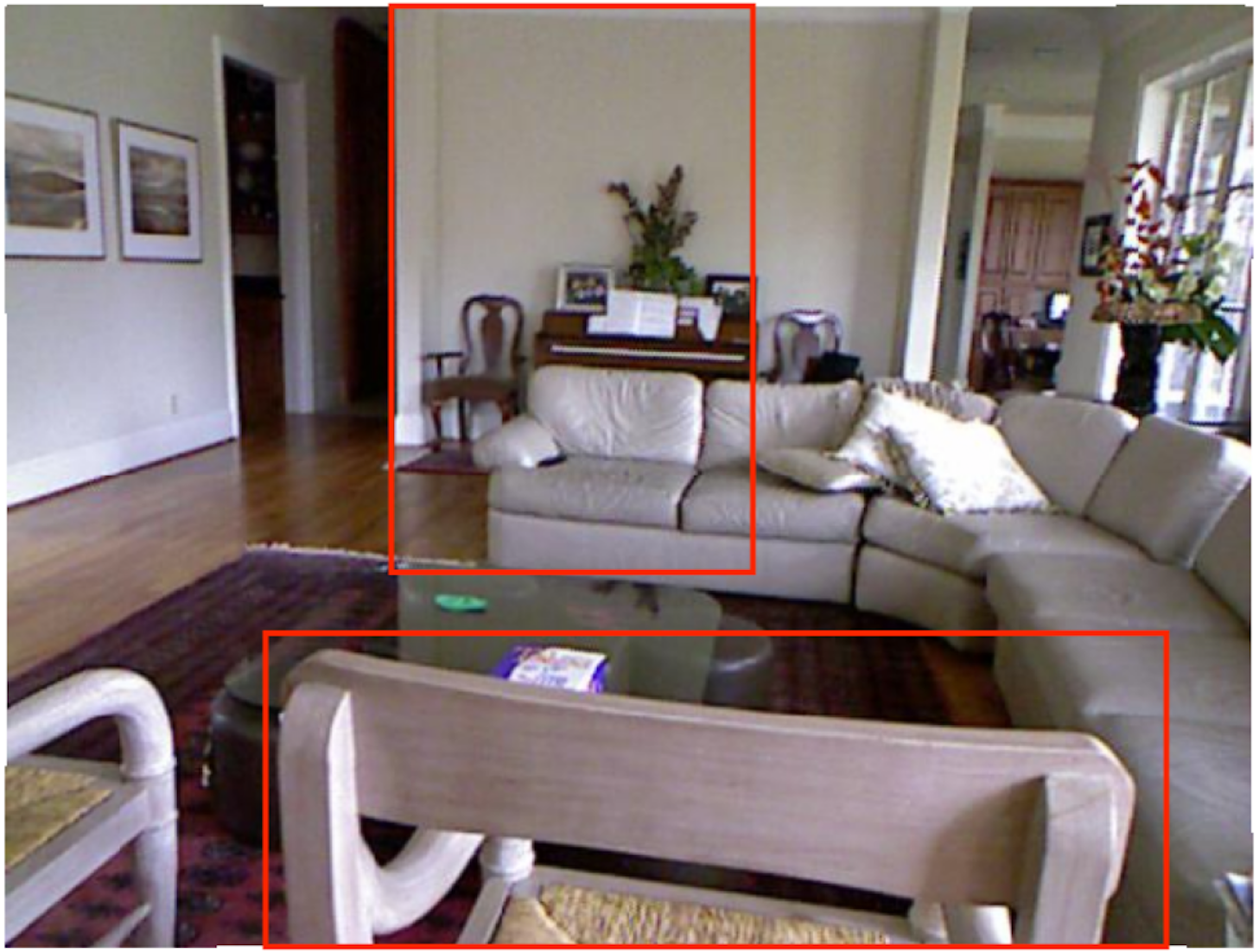}
    \end{minipage}%
    \begin{minipage}[c]{0.19\textwidth}
        \includegraphics[width=\textwidth]{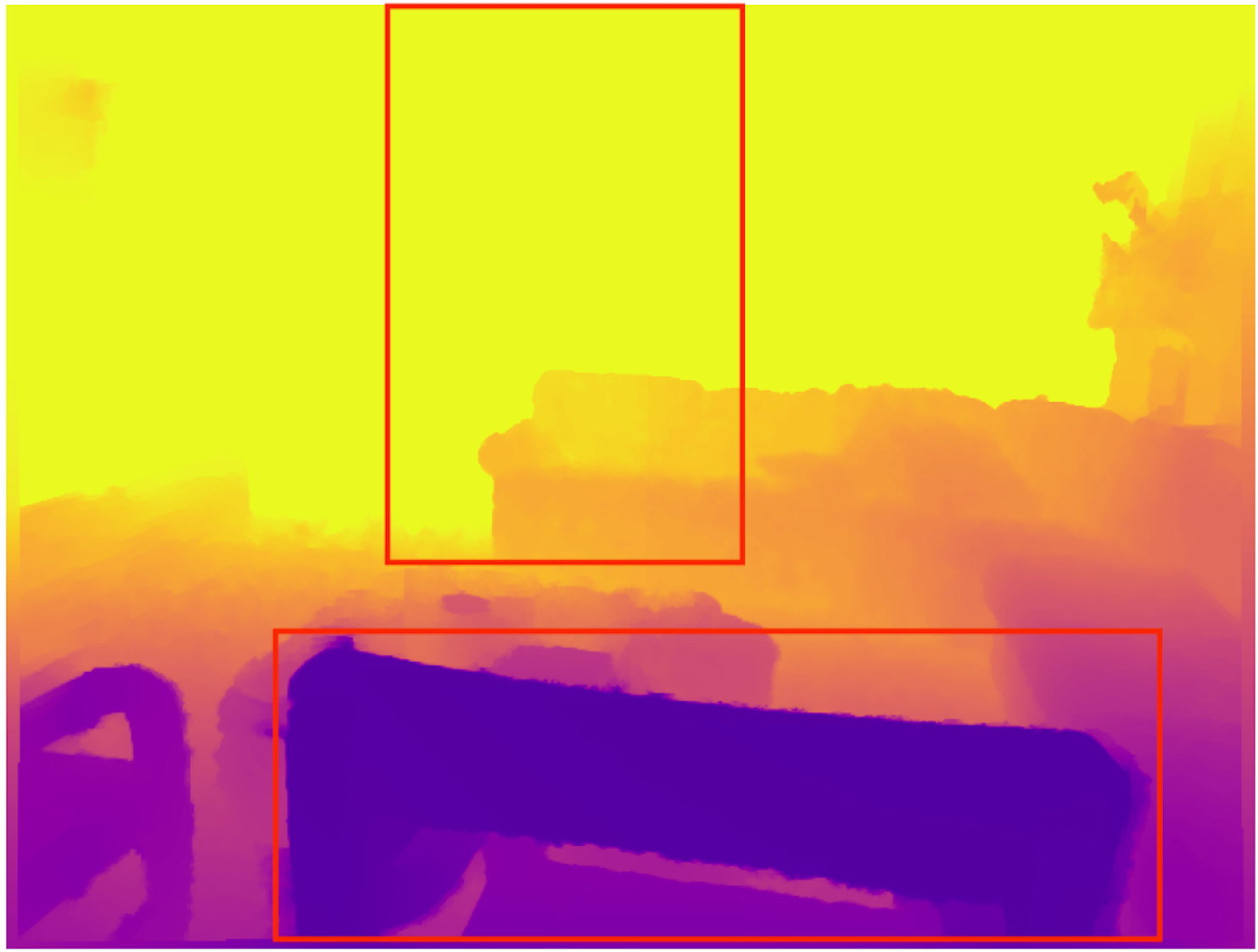}
    \end{minipage}%
    \begin{minipage}[c]{0.19\textwidth}
        \includegraphics[width=\textwidth]{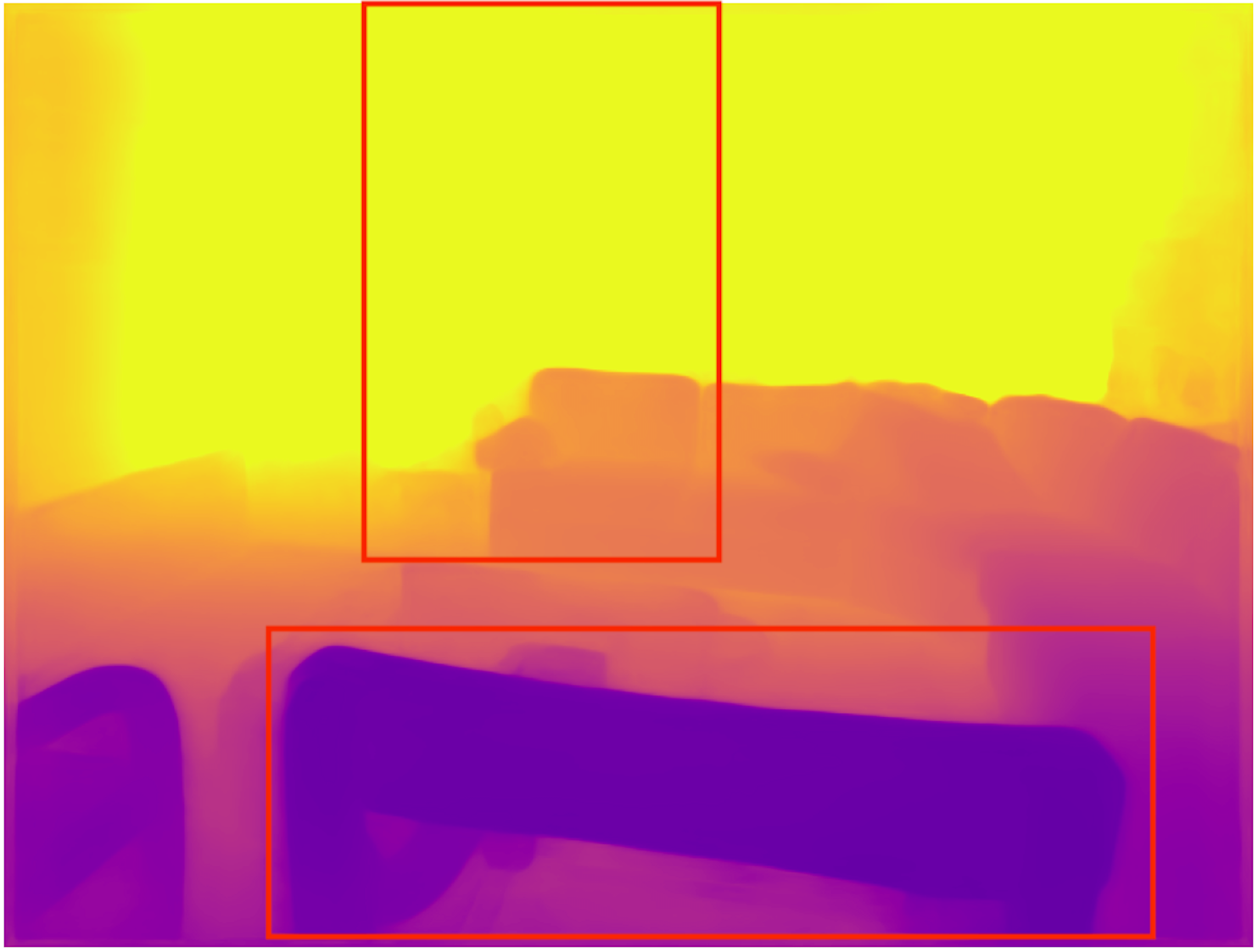}
    \end{minipage}%
    \begin{minipage}[c]{0.19\textwidth}
        \includegraphics[width=\textwidth]{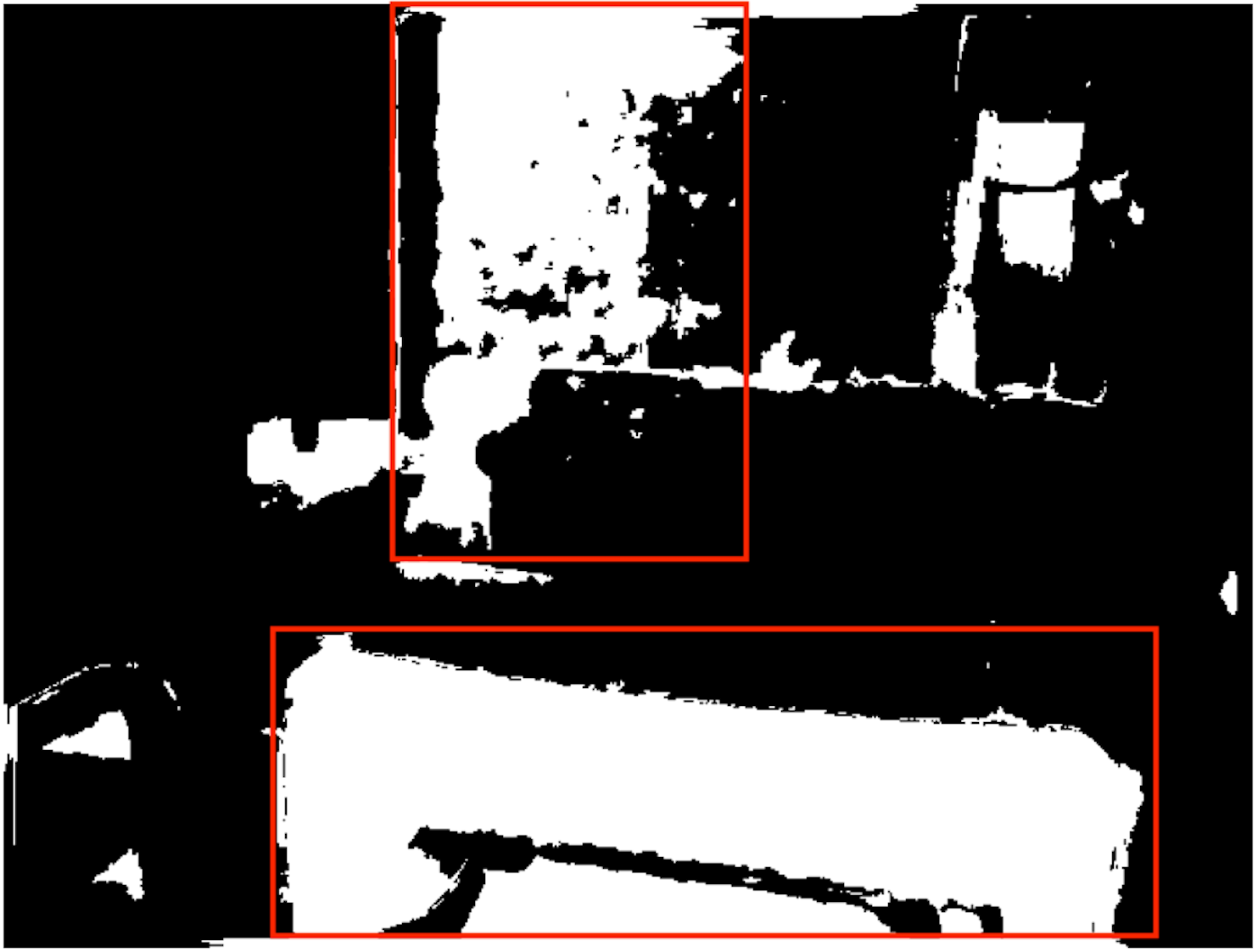}
    \end{minipage}%
    \begin{minipage}[c]{0.19\textwidth}
        \includegraphics[width=\textwidth]{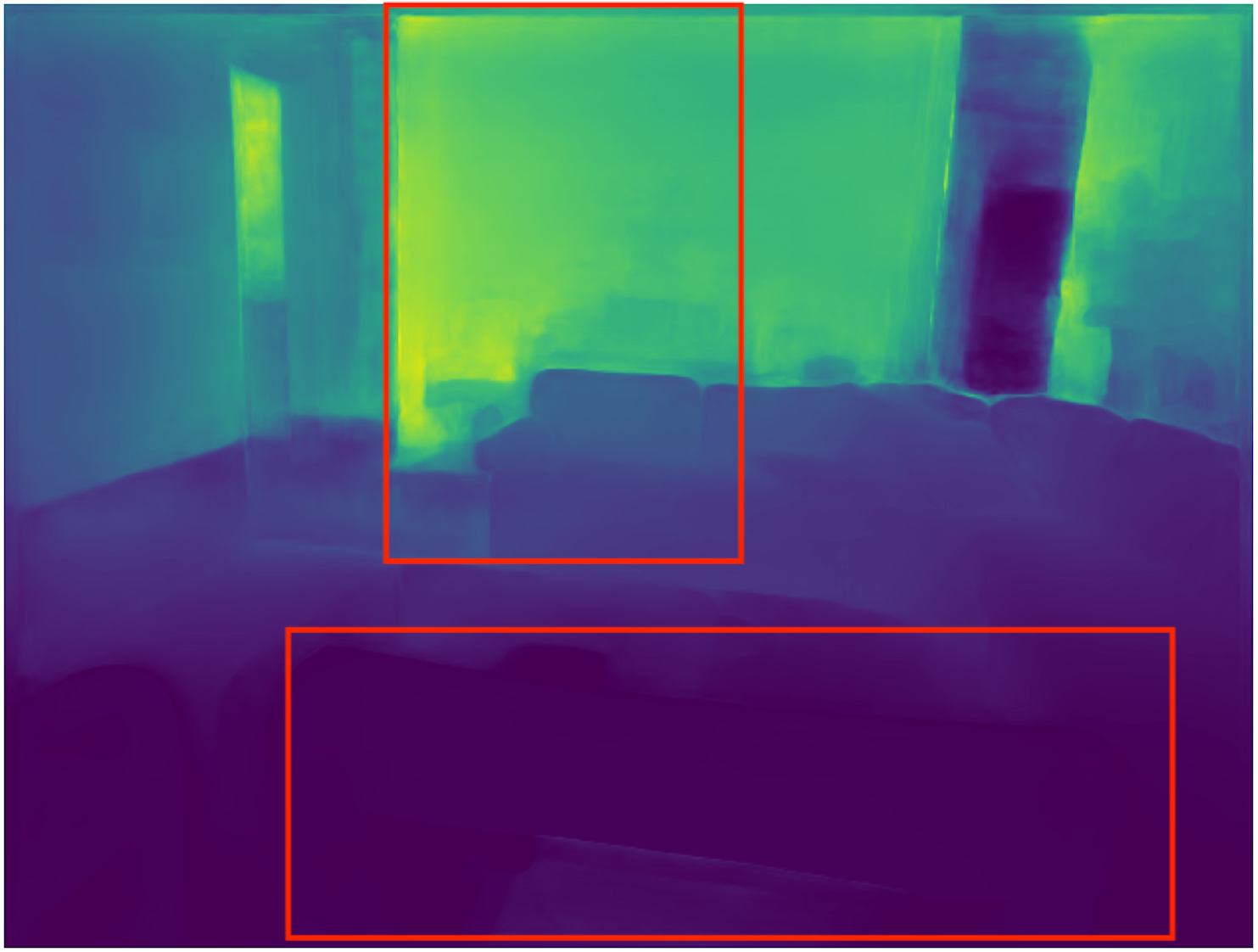}
    \end{minipage}

    \begin{minipage}[c]{0.025\textwidth}
        \centering
        \adjustbox{valign=c}{\rotatebox[origin=c]{90}{MCD (ViT-B)}}
    \end{minipage}%
    \begin{minipage}[c]{0.19\textwidth}
        \includegraphics[width=\textwidth]{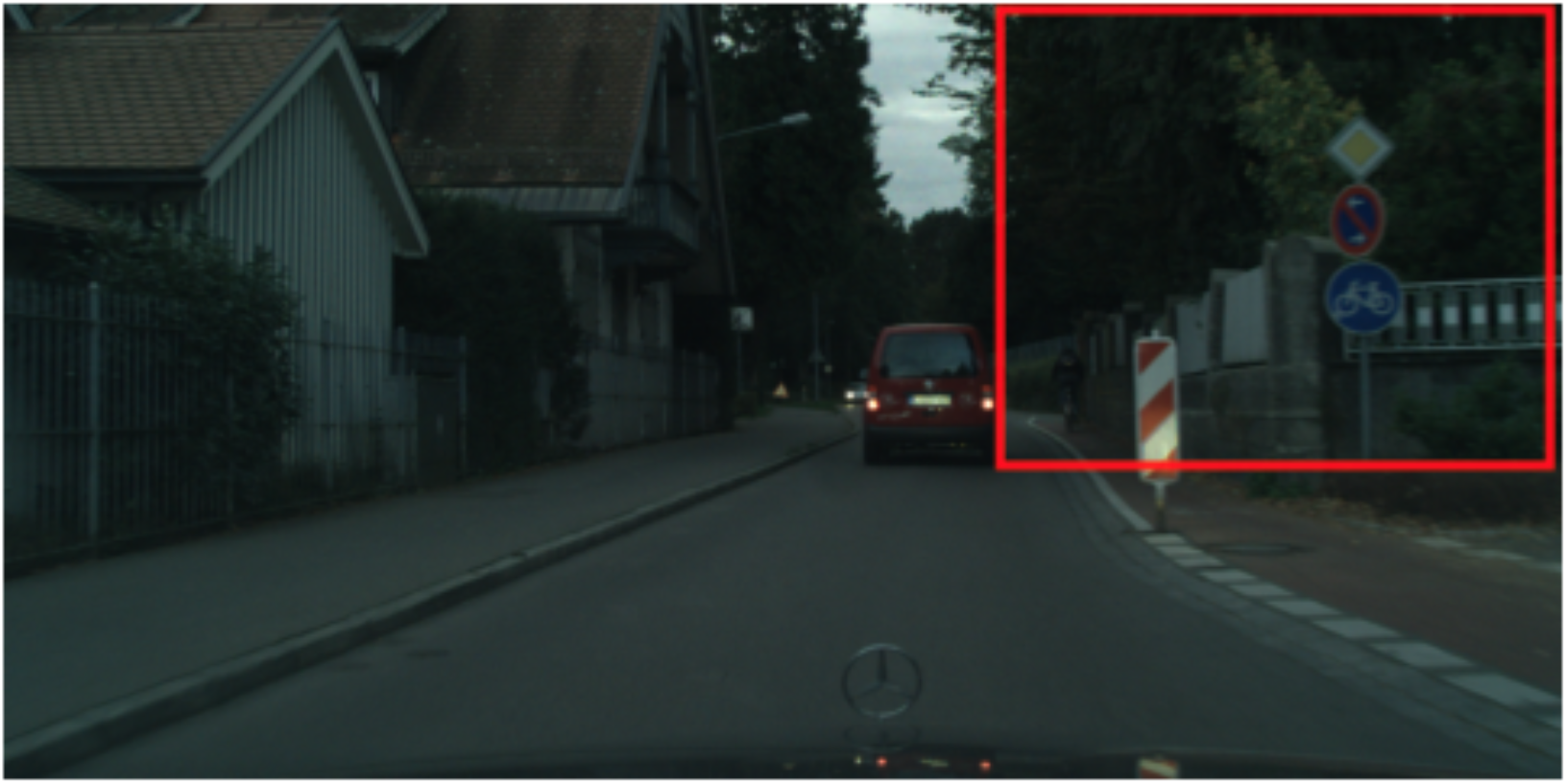}
    \end{minipage}%
    \begin{minipage}[c]{0.19\textwidth}
        \includegraphics[width=\textwidth]{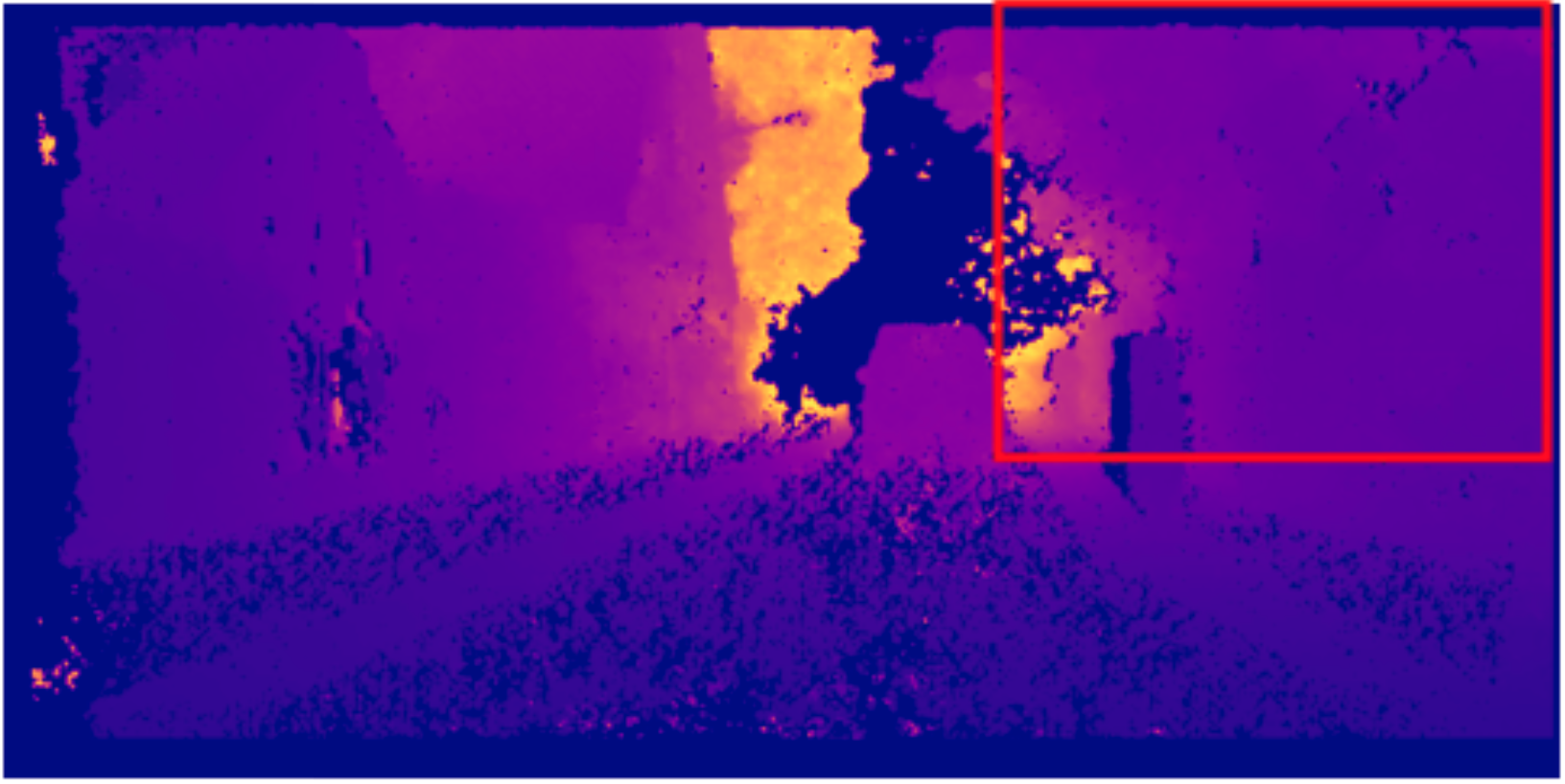}
    \end{minipage}%
    \begin{minipage}[c]{0.19\textwidth}
        \includegraphics[width=\textwidth]{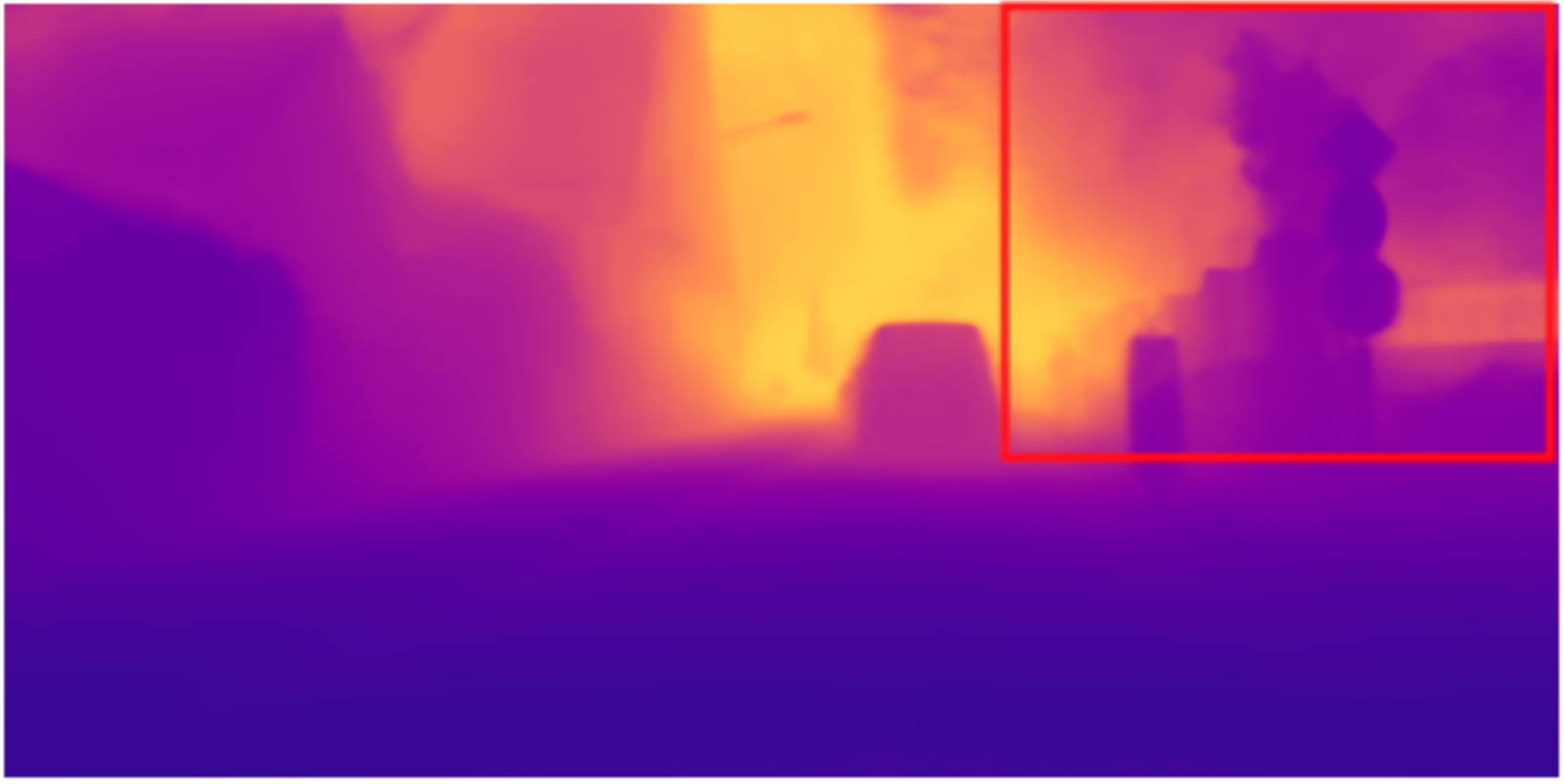}
    \end{minipage}%
    \begin{minipage}[c]{0.19\textwidth}
        \includegraphics[width=\textwidth]{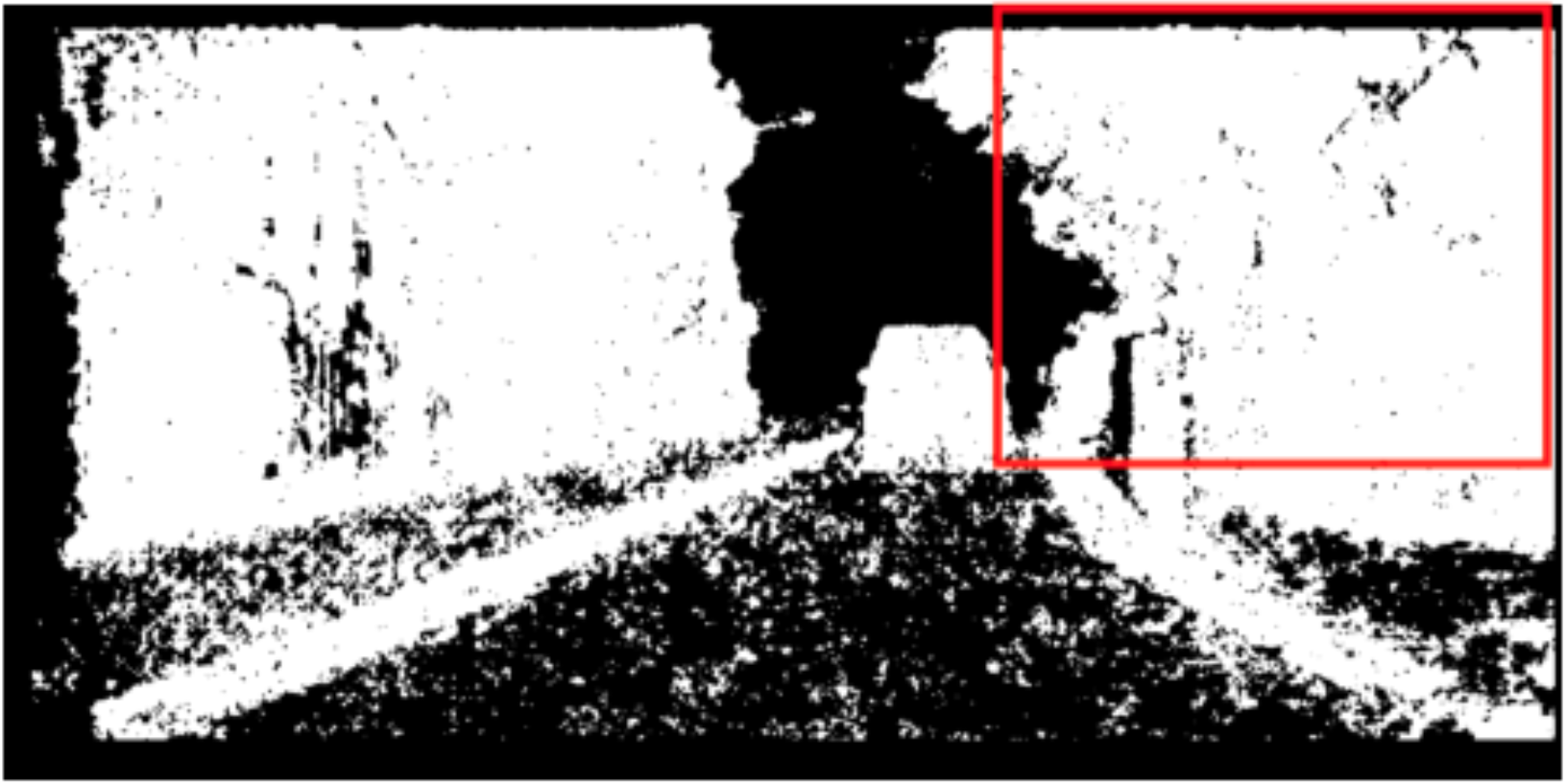}
    \end{minipage}%
    \begin{minipage}[c]{0.19\textwidth}
        \includegraphics[width=\textwidth]{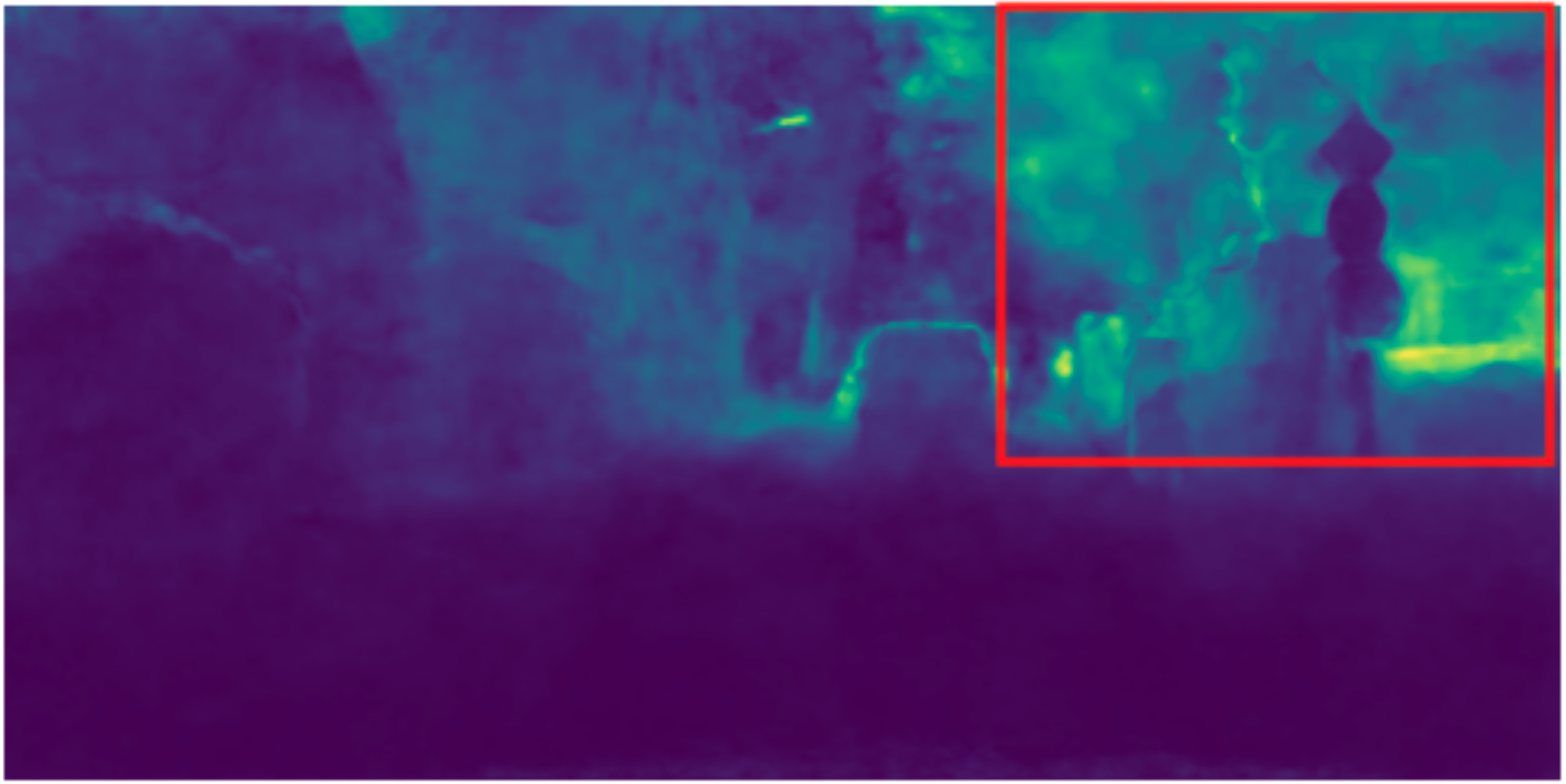}
    \end{minipage}

    \begin{minipage}[c]{0.025\textwidth}
        \centering
        \adjustbox{valign=c}{\rotatebox[origin=c]{90}{SE (ViT-B)}}
    \end{minipage}%
    \begin{minipage}[c]{0.19\textwidth}
        \includegraphics[width=\textwidth]{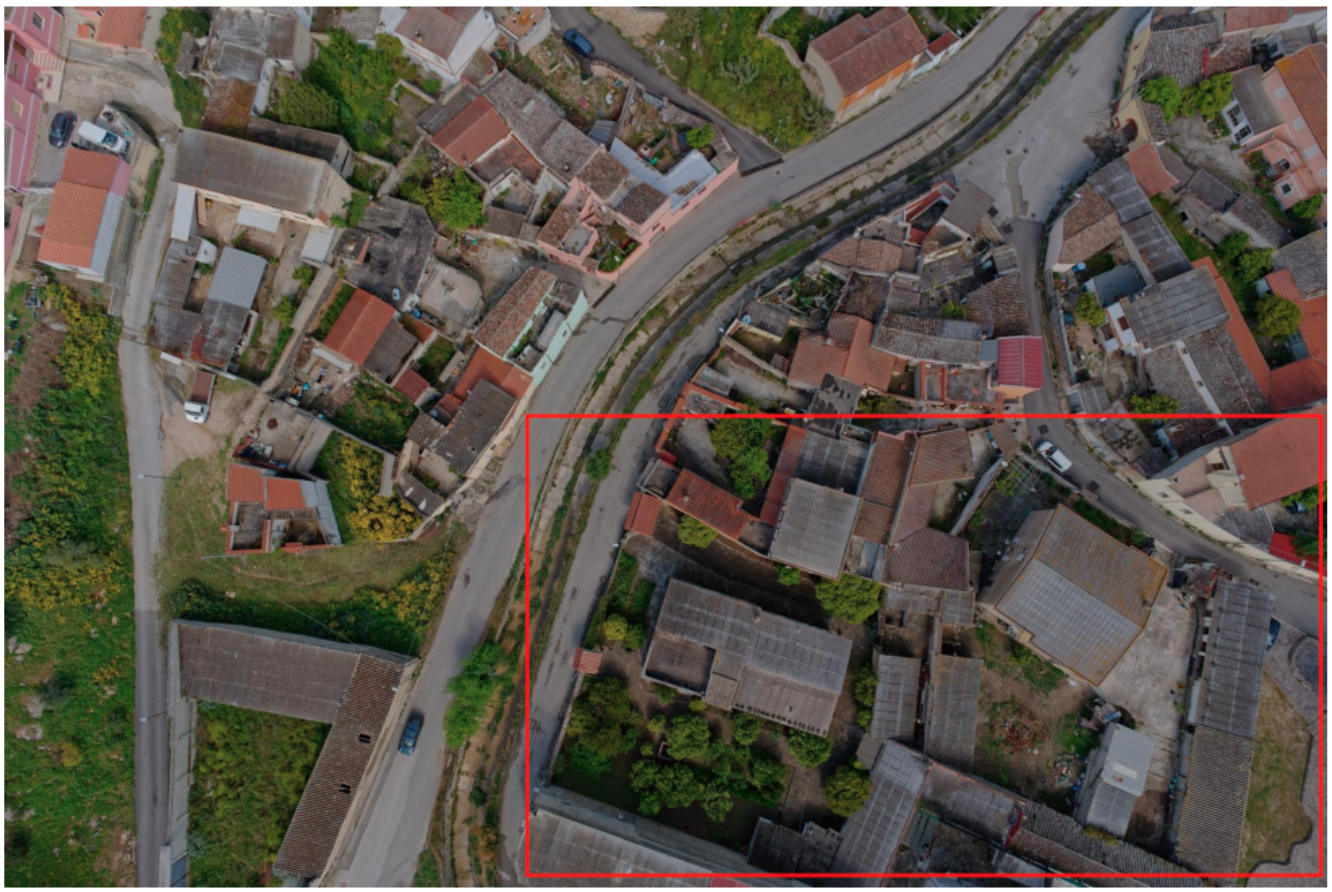}
    \end{minipage}%
    \begin{minipage}[c]{0.19\textwidth}
        \includegraphics[width=\textwidth]{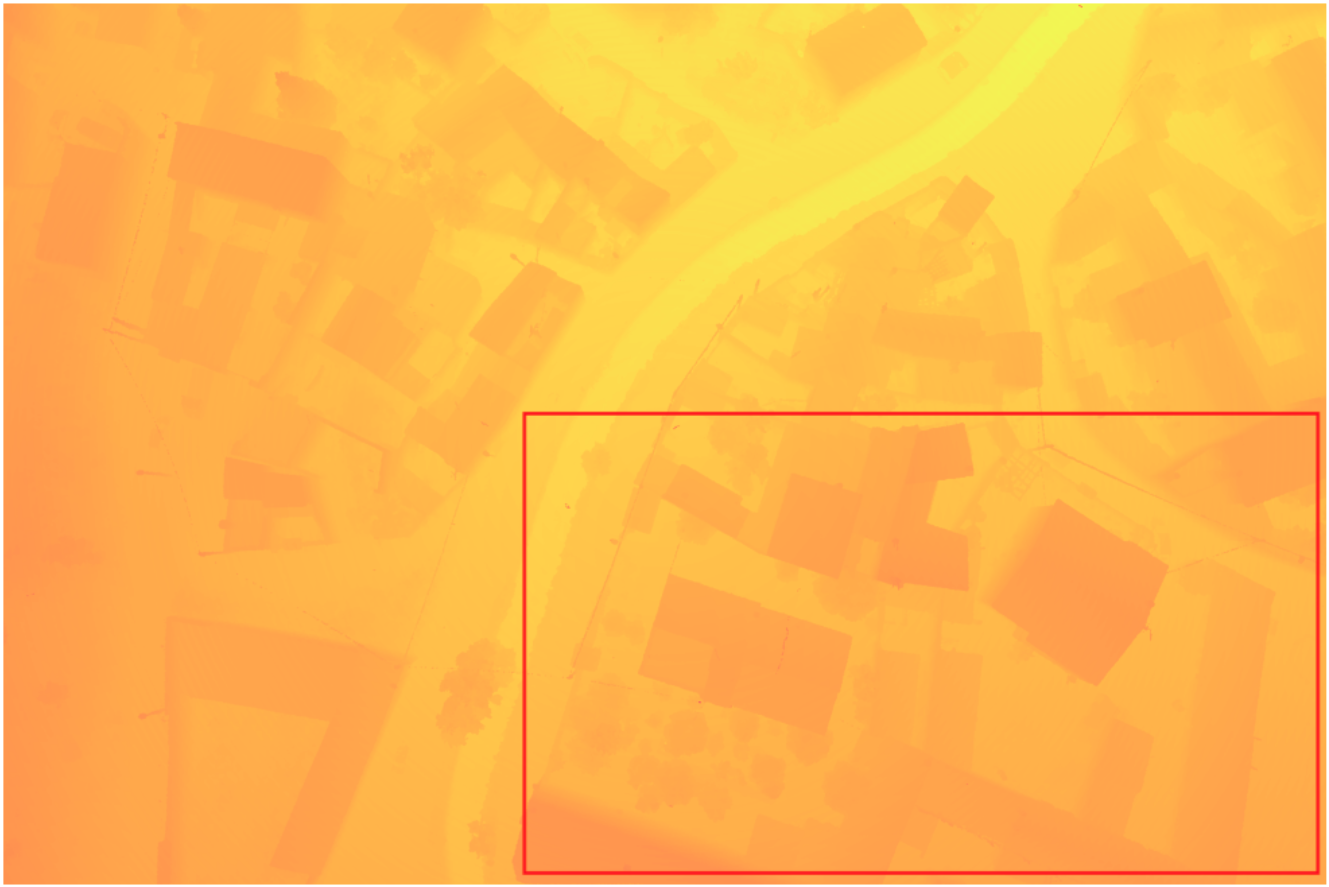}
    \end{minipage}%
    \begin{minipage}[c]{0.19\textwidth}
        \includegraphics[width=\textwidth]{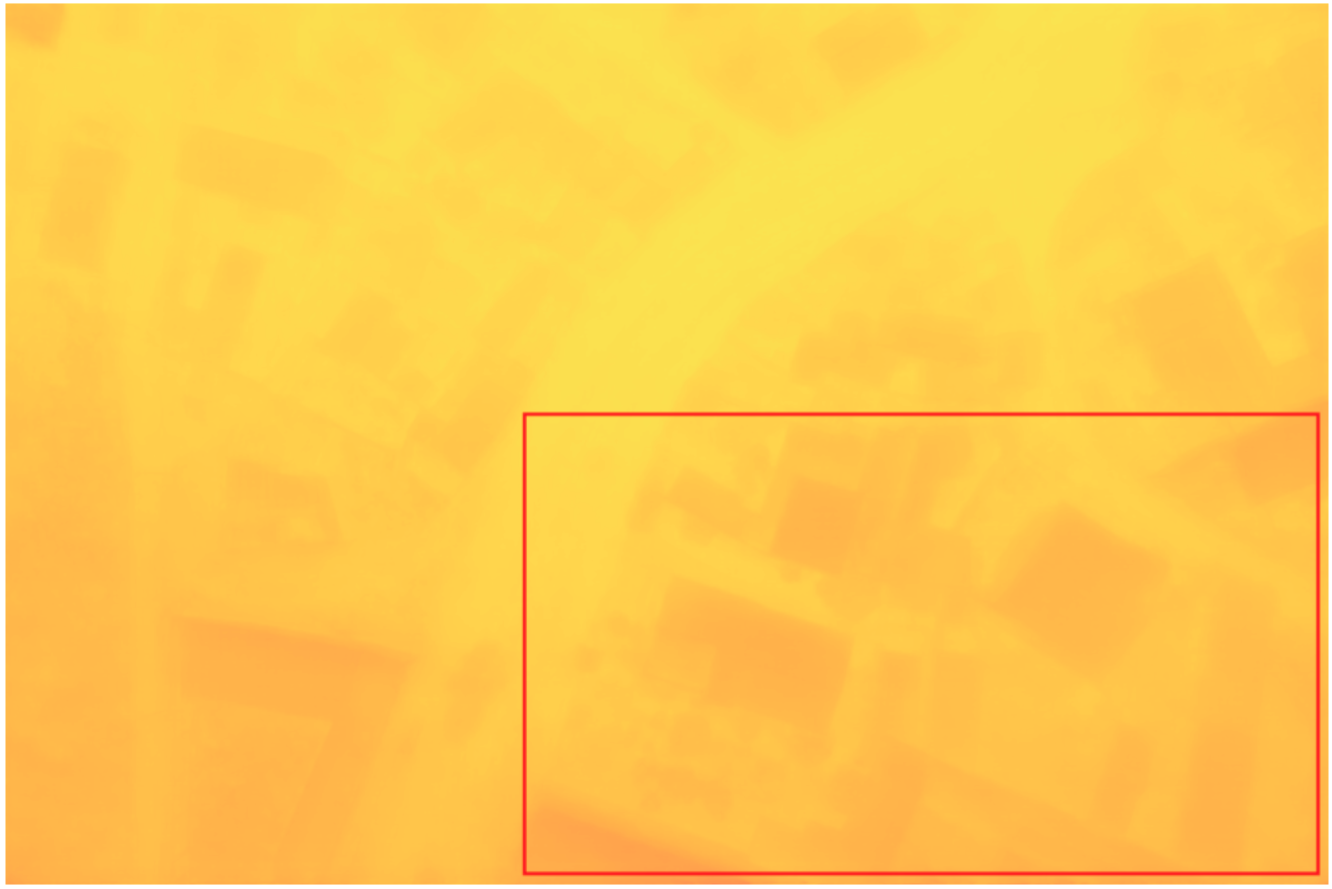}
    \end{minipage}%
    \begin{minipage}[c]{0.19\textwidth}
        \includegraphics[width=\textwidth]{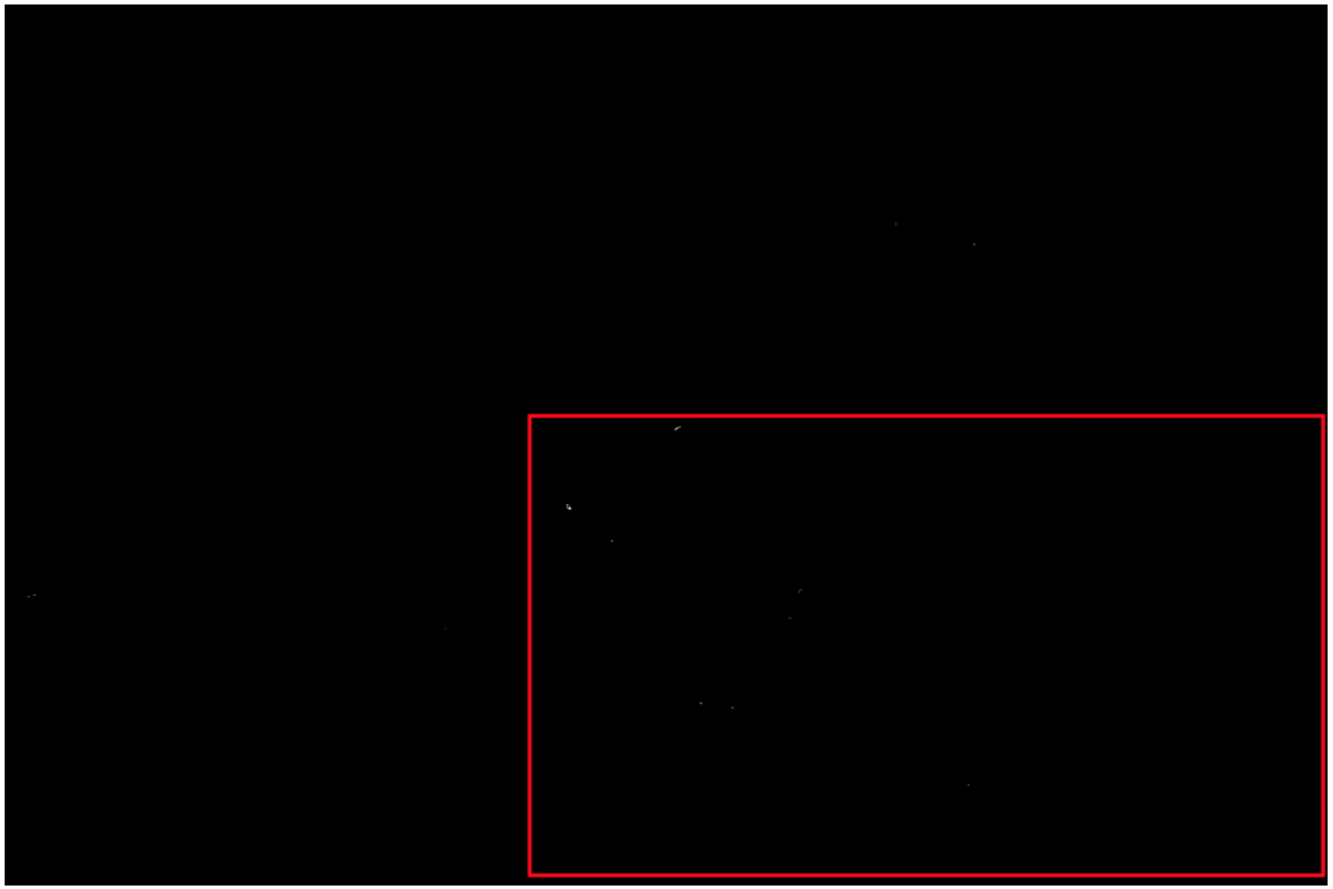}
    \end{minipage}%
    \begin{minipage}[c]{0.19\textwidth}
        \includegraphics[width=\textwidth]{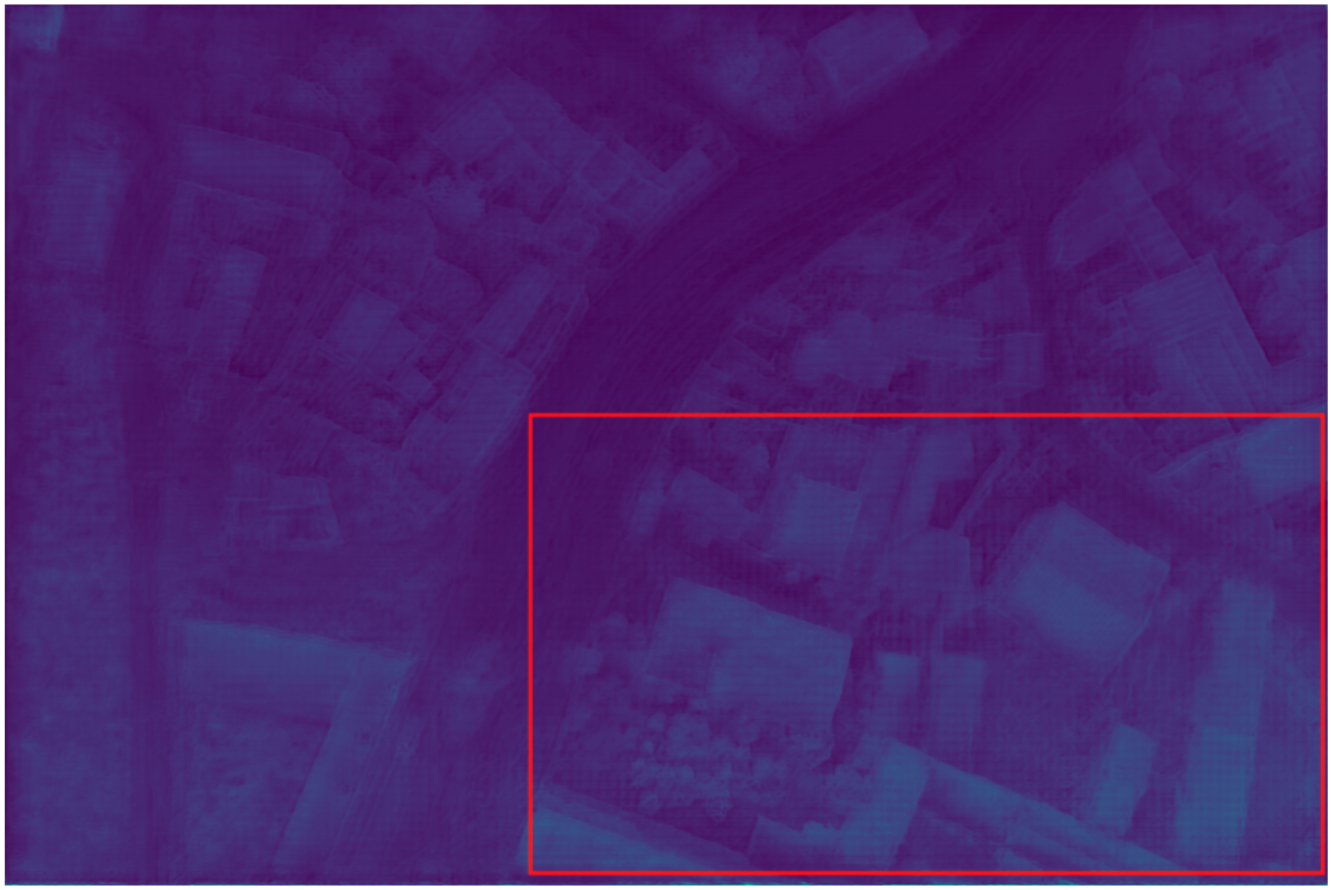}
    \end{minipage}

    \begin{minipage}[c]{0.025\textwidth}
        \centering
        \adjustbox{valign=c}{\rotatebox[origin=c]{90}{LC (ViT-L)}}
    \end{minipage}%
    \begin{minipage}[c]{0.19\textwidth}
        \includegraphics[width=\textwidth]{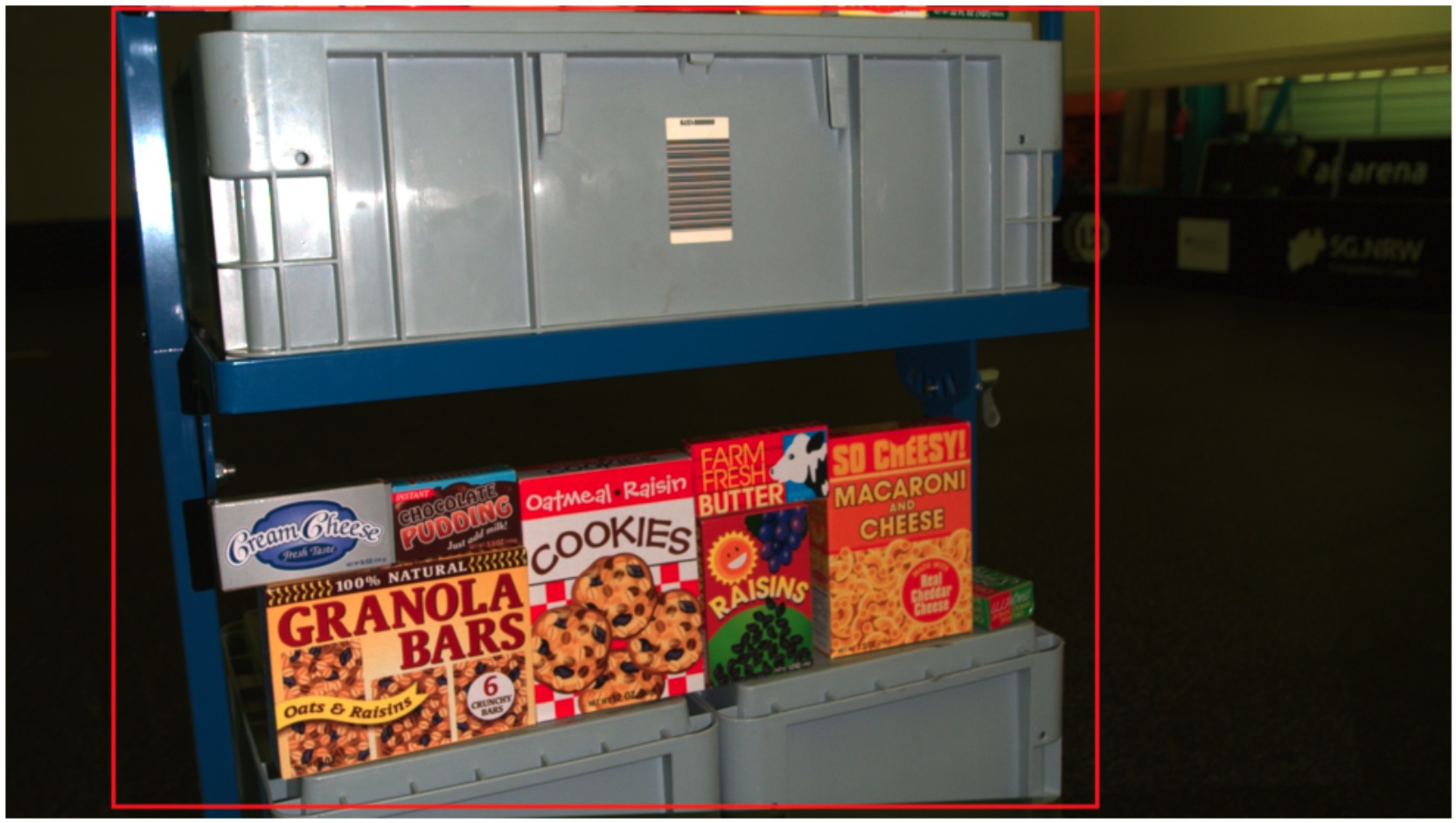}
    \end{minipage}%
    \begin{minipage}[c]{0.19\textwidth}
        \includegraphics[width=\textwidth]{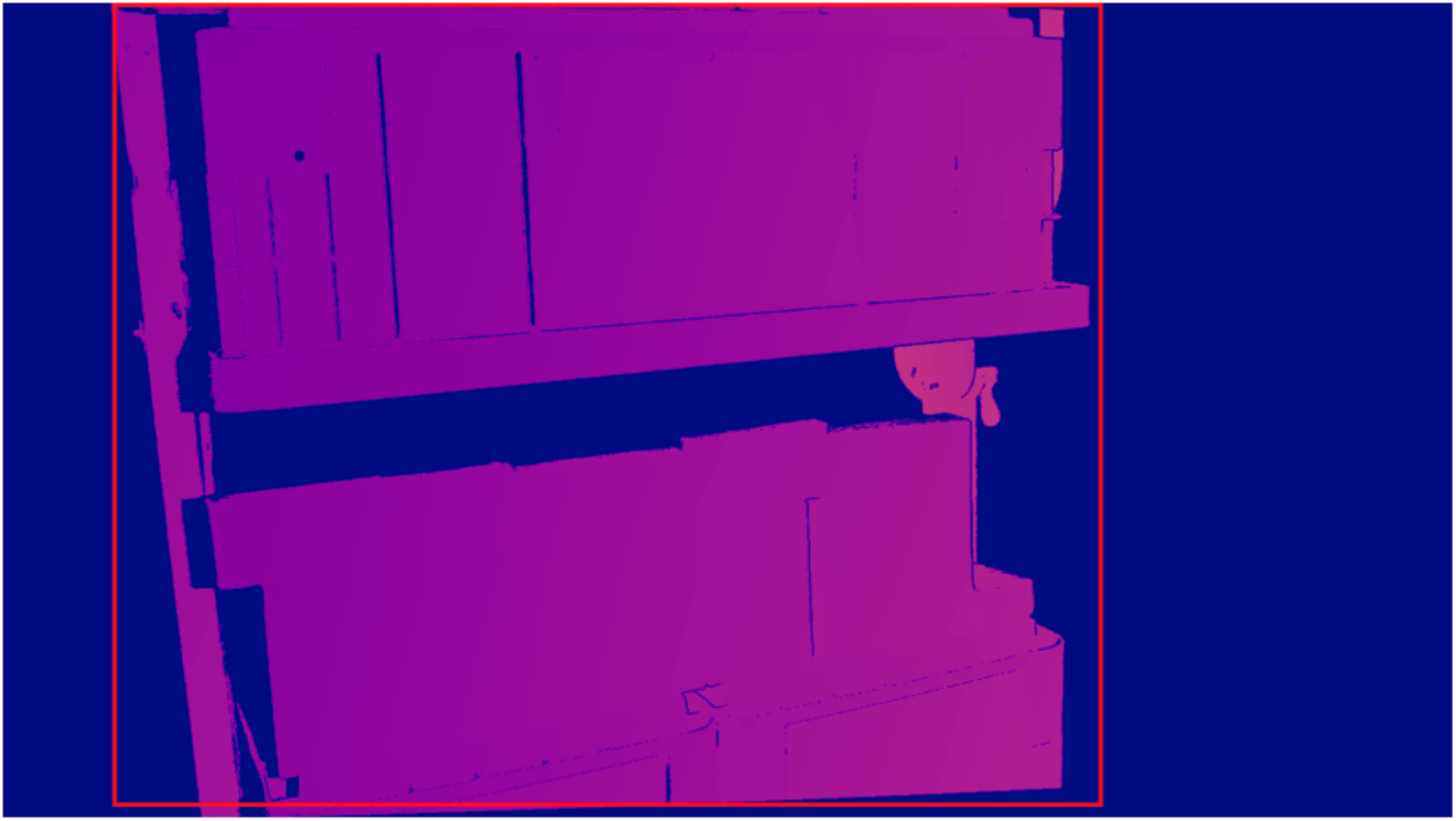}
    \end{minipage}%
    \begin{minipage}[c]{0.19\textwidth}
        \includegraphics[width=\textwidth]{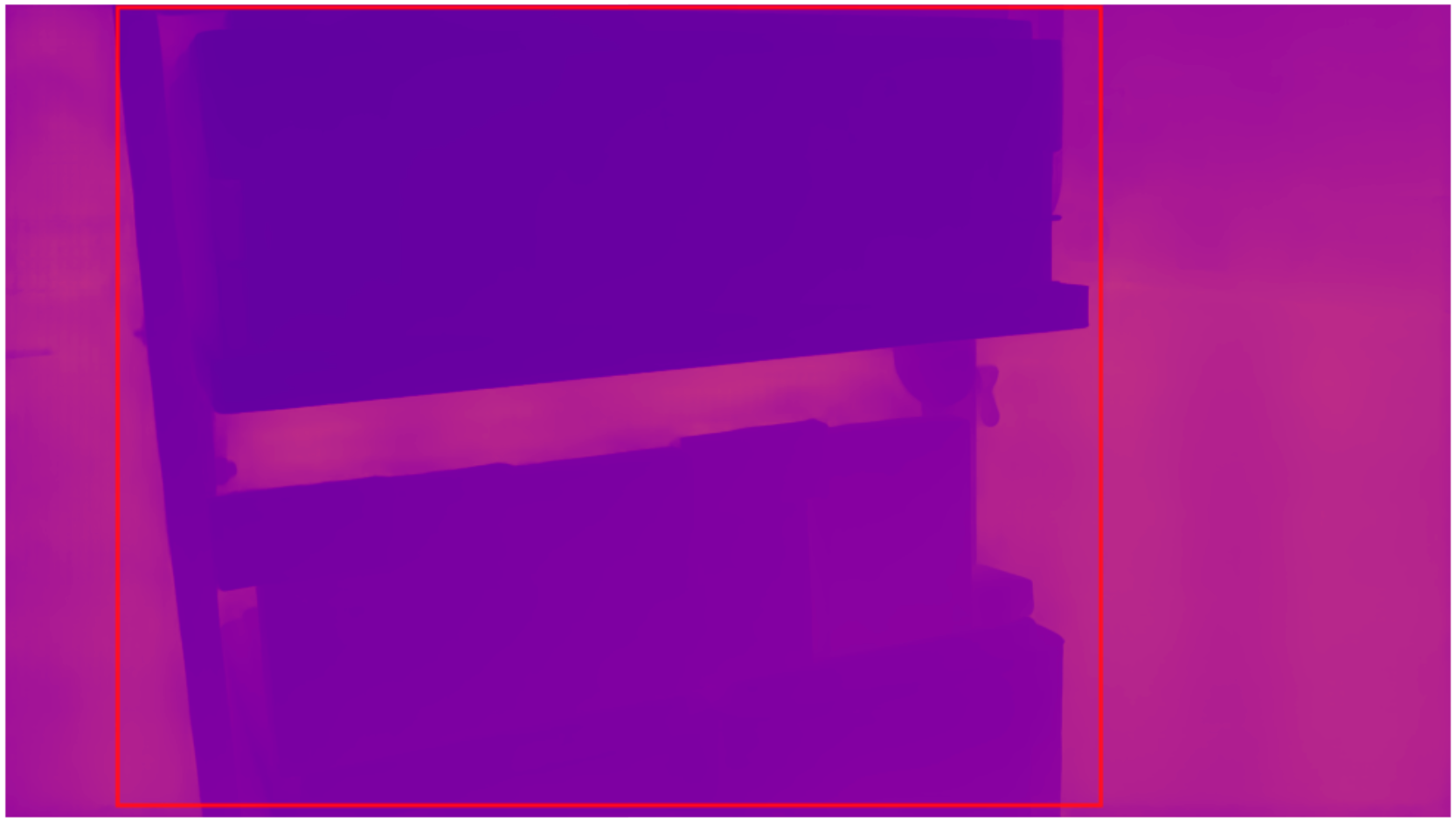}
    \end{minipage}%
    \begin{minipage}[c]{0.19\textwidth}
        \includegraphics[width=\textwidth]{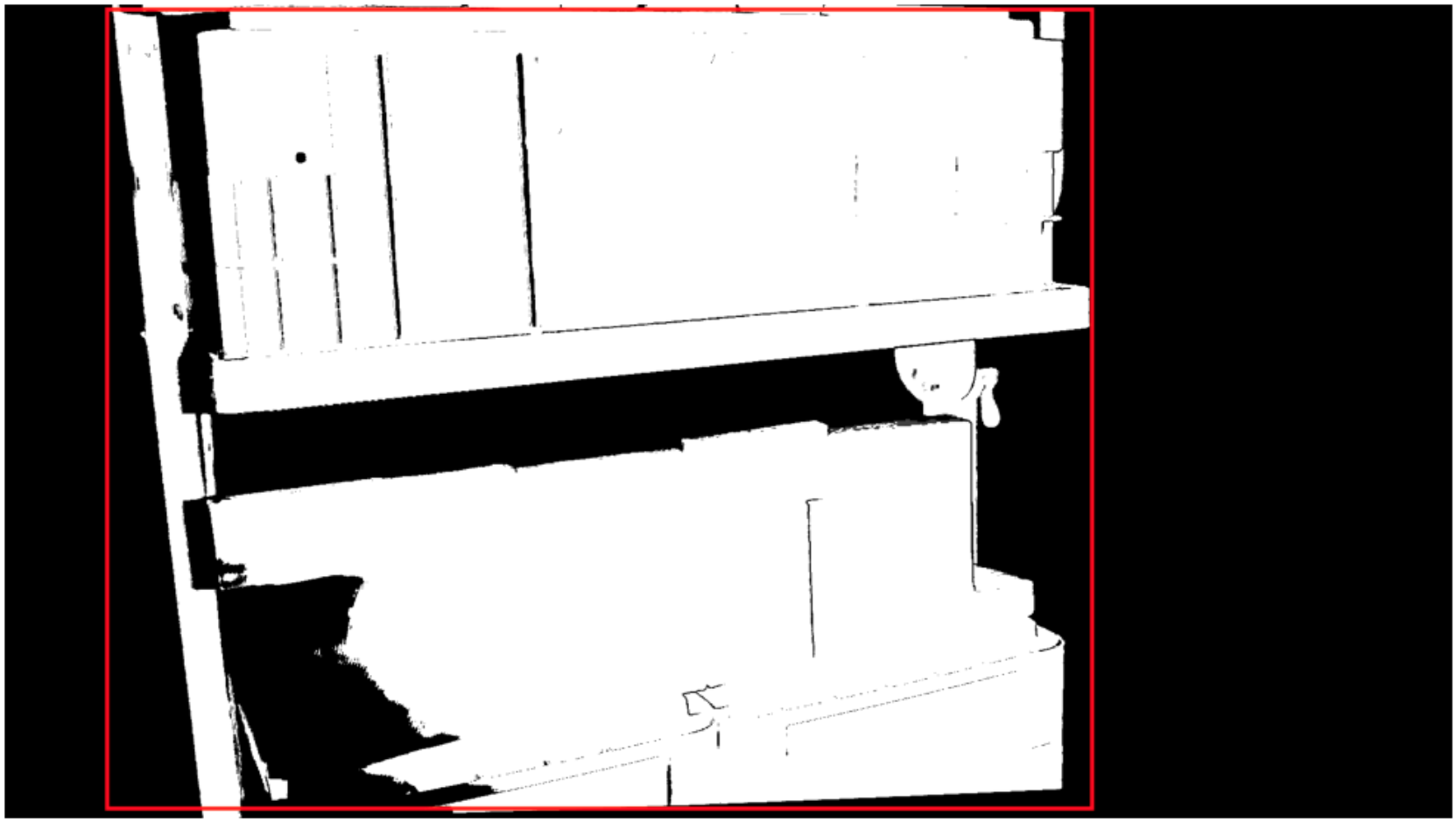}
    \end{minipage}%
    \begin{minipage}[c]{0.19\textwidth}
        \includegraphics[width=\textwidth]{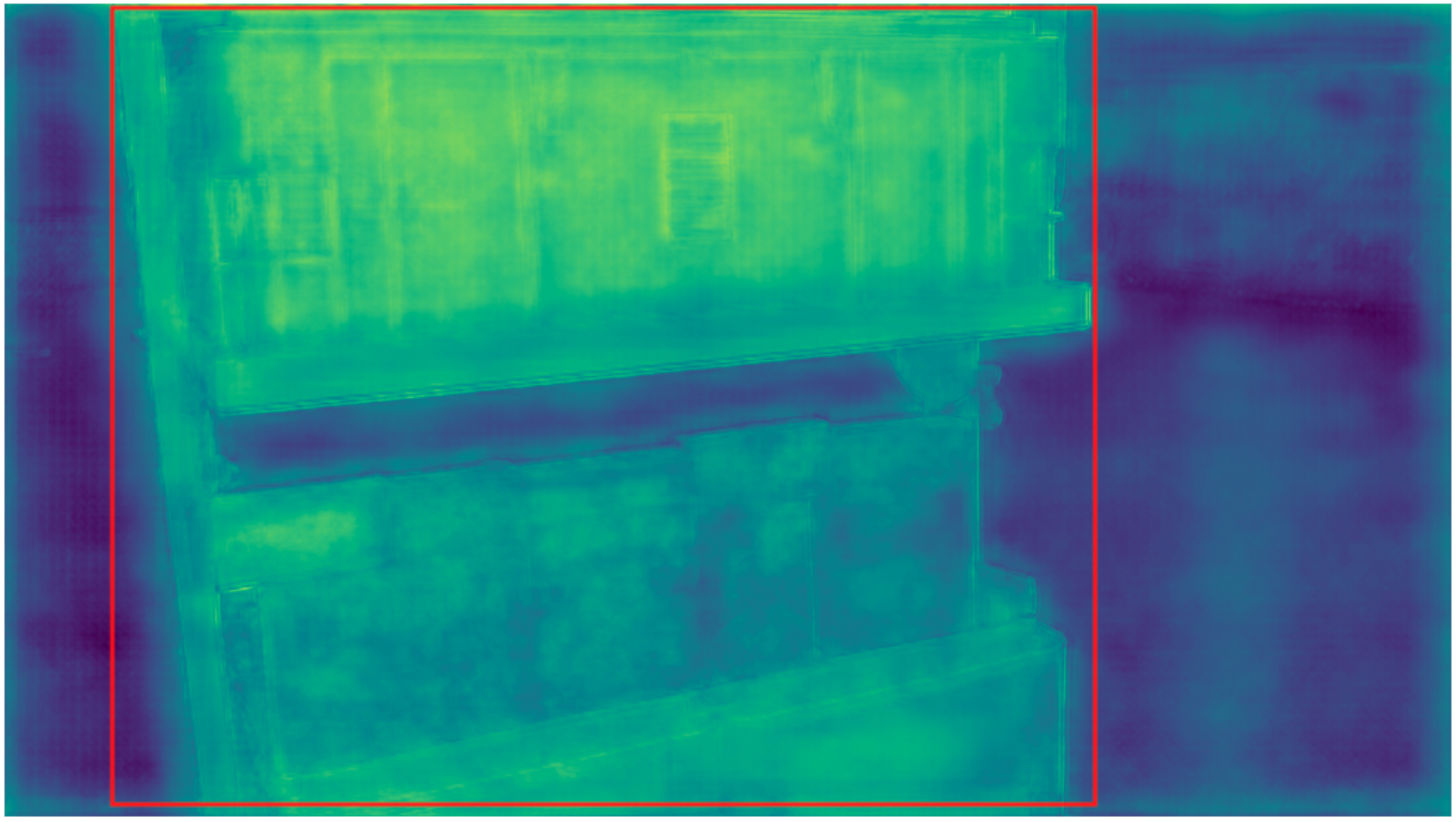}
    \end{minipage}

    \begin{minipage}[c]{0.025\textwidth}
        \phantom{Caption} 
    \end{minipage}%
    \begin{minipage}[c]{0.19\textwidth}
        \caption*{Input Image}
    \end{minipage}%
    \begin{minipage}[c]{0.19\textwidth}
        \caption*{Ground Truth}
    \end{minipage}%
    \begin{minipage}[c]{0.19\textwidth}
        \caption*{Prediction}
    \end{minipage}%
    \begin{minipage}[c]{0.19\textwidth}
        \caption*{$\delta_1$ Error}
    \end{minipage}%
    \begin{minipage}[c]{0.19\textwidth}
        \caption*{Uncertainty}
    \end{minipage}

    \caption{Qualitative examples for indoor \protect\citep{silberman2012indoor}, outdoor \protect\citep{cordts2016cityscapes}, aerial \protect\citep{nex2024usegeo}, and robotics \protect\citep{tyree2022hope} scenarios with varying uncertainty quantification approaches and encoder sizes. Red rectangles are added to highlight interesting areas. Best viewed in color.}
    \label{fig:qualitative_examples}
\end{figure*}

We provide handpicked qualitative examples of four different UQ methods with varying encoder sizes for all four datasets in Figure \ref{fig:qualitative_examples}, highlighting the potential for foundation model uncertainty in metric MDE.

\textbf{NYUv2.}
The first row displays the GNLL (ViT-S) results, demonstrating high prediction quality overall. Uncertainty is notably elevated around object boundaries and the two open doors in the background, suggesting these regions are likely edge cases. This likely stems from the model's limited exposure to such depth ranges during training since we limit the maximum depth to just 5m, which is common practice on NYUv2.

Additionally, the second row presents results from TTA (ViT-S), revealing mixed results in terms of prediction and uncertainty quality. While the model assigns high uncertainties to the background, which includes some erroneous predictions, it fails to recognize its substantial prediction error on the backrest of the chair at the bottom of the image. Based on this qualitative comparison, GNLL seems to provide more meaningful uncertainties, corroborating the quantitative findings of Section \ref{sec: quantitative_evaluation}.

\textbf{Cityscapes.}
In the third row, the MCD (ViT-B) predictions exhibit multiple errors, particularly in the top right corner. However, the predicted uncertainties in this area show a correlation between uncertainty and challenging regions, reinforcing the model's awareness of its limitations. 

\textbf{UseGeo.} 
The fourth row shows SE (ViT-B) results, where relative depth predictions are plausible despite reduced accuracy in absolute terms. Across the entire image, especially in the bottom right, uncertainties are heightened for the buildings, emphasizing that the model is aware of key areas where depth errors are most likely. 

\textbf{HOPE.}
The fifth row presents LC (ViT-L) results, where the model struggles with the absolute depth of the large foreground object. At the same time, the entire object is highlighted by high uncertainty, reflecting the model's strong awareness of its own prediction error in this case.

\section{Conclusion}
Motivated by the need to bridge the gap between cutting-edge research and the safe deployment of MDE models in real-world applications, we conducted a comprehensive evaluation of multiple UQ methods in conjunction with the state-of-the-art DepthAnythingV2 foundation model. Our evaluation covered five different UQ approaches -- Learned Confidence, Gaussian Negative Log-Likelihood, MC Dropout, Sub-Ensembles, and Test-Time Augmentation -- and was carried out across four diverse dataset, covering various domains relevant to real-world applications: NYUv2, Cityscapes, UseGeo, and HOPE.

Our findings highlight fine-tuning with GNLL as the most promising option, consistently delivering high-quality uncertainty estimates while maintaining depth performance comparable to the baseline. Its computational efficiency, which matches that of the baseline, further underscores its practical suitability for deployment.

This study emphasizes the importance and feasibility of integrating UQ into machine vision models, demonstrating that achieving reliable uncertainty estimates need not come at the expense of predictive performance or computational complexity. By addressing this critical aspect, we aim to inspire future research that prioritizes not only performance but also explainability through uncertainty awareness, fostering the development of safer and more reliable models for not only MDE but also other real-world tasks, such as semantic segmentation or pose estimation. 

\textbf{Acknowledgements.}\\
The authors acknowledge support by the state of Baden-Württemberg through bwHPC.\\
This work is supported by the Helmholtz Association Initiative and Networking Fund on the HAICORE@KIT partition.\\
The work is partially supported by the Office of Naval Research (N00014-23-1-2670).\\

{
\begin{spacing}{1.17}
    \normalsize
    \bibliography{ISPRSguidelines_authors} 
\end{spacing}
}

\end{document}